\documentclass[journal,usenames,dvipsnames]{IEEEtran}

\usepackage{microtype}
\usepackage{graphicx}
\usepackage{booktabs} % for professional tables
\usepackage{graphicx}
\usepackage{svg}
\usepackage{amssymb}
\usepackage{fancyhdr,graphicx,amsmath}
\usepackage[ruled,vlined,algo2e]{algorithm2e}
\usepackage{xcolor}
\usepackage{silence}
\usepackage{mathtools}
\usepackage{algorithm}
\usepackage{makecell}

\makeatletter
\newcommand*{\rom}[1]{\expandafter\@slowromancap\romannumeral #1@}
\makeatother
\usepackage{ifpdf}
\usepackage{cite}
\usepackage{amsmath}
\usepackage{algorithmic}
\usepackage{array}
\usepackage[caption=false,font=normalsize,labelfont=sf,textfont=sf]{subfig}
\usepackage{stfloats}
\usepackage{url}

\makeatletter
\def\thickhline{%
  \noalign{\ifnum0=`}\fi\hrule \@height \thickarrayrulewidth \futurelet
   \reserved@a\@xthickhline}
\def\@xthickhline{\ifx\reserved@a\thickhline
               \vskip\doublerulesep
               \vskip-\thickarrayrulewidth
             \fi
      \ifnum0=`{\fi}}
\makeatother

\newlength{\thickarrayrulewidth}
\usepackage{booktabs,makecell,multirow,tabularx}
\newcolumntype{L}{>{\RaggedRight\arraybackslash}X}
\newcolumntype{C}{>{\Centering\arraybackslash}X}

\newcommand\mc[1]{\multicolumn{3}{c}{#1}}
\usepackage{ragged2e}

\newcommand{\revised}[1]{\textcolor{black}{#1}}

\newcommand{\giannis}[1]{\textcolor{blue}{#1}}

\usepackage[inline]{enumitem}
\newcommand{\etal}{et al.\ }  % no \textit, since cite doesn't do it either

\newcommand{\ie}{i.e., }

\newcommand{\figref}[1]{Fig.~\ref{#1}}    % within sentence
\newcommand{\Figref}[1]{Figure~\ref{#1}}  % start of sentence
\newcommand{\tabref}[1]{Table~\ref{#1}}
\newcommand{\algref}[1]{Algorithm~\ref{#1}}

\newcommand{\secref}[1]{Section~\ref{#1}}

\usepackage{tikz,xcolor,hyperref}

\definecolor{lime}{HTML}{A6CE39}
\DeclareRobustCommand{\orcidicon}{
    \begin{tikzpicture}
    \draw[lime, fill=lime] (0,0) 
    circle [radius=0.16] 
    node[white] {{\fontfamily{qag}\selectfont \tiny ID}};
    \draw[white, fill=white] (-0.0625,0.095) 
    circle [radius=0.007];
    \end{tikzpicture}
    \hspace{-2mm}
}

\foreach \x in {A, ..., Z}{\expandafter\xdef\csname orcid\x\endcsname{\noexpand\href{https://orcid.org/\csname orcidauthor\x\endcsname}
            {\noexpand\orcidicon}}
}
\begin{document}
%
% paper title
% Titles are generally capitalized except for words such as a, an, and, as,
% at, but, by, for, in, nor, of, on, or, the, to and up, which are usually
% not capitalized unless they are the first or last word of the title.
% Linebreaks \\ can be used within to get better formatting as desired.
% Do not put math or special symbols in the title.
\title{Learned Gradient Compression \\ for Distributed Deep Learning}

\author{Lusine~Abrahamyan\orcidD{}, Yiming Chen\orcidC{}, Giannis~Bekoulis\orcidB{}, and~Nikos~Deligiannis\orcidA{},~\IEEEmembership{Member,~IEEE}
        % <-this % stops a space

\thanks{L. Abrahamyan, Y. Chen, G. Bekoulis, and N. Deligiannis are with Vrije Universiteit Brussel, Pleinlaan 2, B-1050 Brussels, Belgium and also with imec, Kapeldreef 75, B-3001 Leuven, Belgium. (e-mail: \{alusine, ychen, gbekouli,ndeligia\}@etrovub.be).}% <-this % stops a space
}

\maketitle

\begin{abstract}
    Training deep neural networks on large datasets containing high-dimensional data  requires a large amount of computation. A solution to this problem is data-parallel distributed training, where a model is replicated into several computational nodes that have access to different chunks of the data. This approach, however, entails high communication rates and latency because of the computed gradients that need to be shared among nodes at every iteration. The problem becomes more pronounced in the case that there is wireless communication between the nodes (\ie due to the limited network bandwidth).
    To address this problem, various compression methods have been proposed including sparsification, quantization, and entropy encoding of the gradients. Existing methods leverage the \textit{intra-node} information redundancy, that is, they compress gradients at each node \textit{independently}. In contrast, we advocate that the gradients across the nodes are correlated and propose methods to leverage this \textit{inter-node} redundancy to improve compression efficiency. 
    Depending on the node communication protocol (\textit{parameter server} or \textit{ring-allreduce}), we propose two instances of the LGC approach that we coin \textit{Learned Gradient Compression} (LGC).
    Our methods exploit an autoencoder (\ie trained during the first stages of the distributed training) to capture the common information that exists in the gradients of the distributed nodes.
    To constrain the nodes' computational complexity, the autoencoder is realized with a lightweight neural network.
    We have tested our LGC methods on the image classification and semantic segmentation tasks using different convolutional neural networks (ResNet50, ResNet101, PSPNet) and multiple datasets (ImageNet, Cifar10, CamVid). 
   The ResNet101 model trained for image classification on Cifar10 achieved significant compression rate reductions with the accuracy of $93.57\%$, which is lower than the baseline distributed training with uncompressed gradients only by $0.18\%$. The rate of the model is reduced by 8095$\times$ and 8$\times$ compared to the baseline and the state-of-the-art deep gradient compression (DGC) method, respectively.
\end{abstract}

\begin{IEEEkeywords}
Deep learning, data-parallel distributed training, gradient compression, autoencoders.
\end{IEEEkeywords}

\IEEEpeerreviewmaketitle

%%%%%%%%%%%%%%%%%%%%%%
\section{Introduction}
\label{sec:introduction}
%%%%%%%%%%%%%%%%%%%%%%
    Recent successful results in the field of artificial intelligence~(AI) are achieved with deep learning models that contain a large number of parameters and are trained using a massive amount of data. For example, FixResNeXt-101 32x48d~\cite{fixresnet}, a state-of-the-art model for image classification, contains approximately 800 million parameters, and BERT (Bidirectional Encoder Representations from Transformers)~\cite{bert}, a recent model for natural language processing, contains 110 million parameters. Training such deep networks in a single machine (given a fixed set of hyperparameters) can take weeks. 
 
    An answer to this problem is to perform the training of such networks in a number of computing nodes in parallel.
    Two parallelization approaches have been exploited in the literature:
    \begin{enumerate*}[label=(\arabic*)]
    \item ~\textit{model-parallel} distributed training, where the different nodes train different parts of the model; and
    \item ~\textit{data-parallel} distributed training, where each node has a replica (\ie a copy) of the model and access to a chunk of the data. 
    \end{enumerate*}
    In both distributed training approaches, there is a communication and latency overhead due to the transmission of information from the one node to the other. In the model-parallel distributed training approach, the data that need to be transferred consist of the activation values of a certain layer in the model. The transmission overhead in this case is typically small. In the data-parallel training approach, however, the calculated gradients of a model that need to be transferred can reach hundreds of megabytes (MBs) per iteration. In this paper, we focus on data-parallel distributed training and propose a new framework to reduce the overhead related to the gradient data transfer.

    Most of the methods that aim to reduce the gradient communication bandwidth consider that the nodes exchange gradients in a synchronous manner. Nevertheless, the study in~\cite{Dean} proposed Downpour Stochastic Gradient Descent, where training is asynchronous in two ways: 
    \begin{enumerate*}[label=(\textit{\roman*)}]
    \item model replicas (\ie copies of the same model) update their gradients at different time instances (asynchronously), and
    \item the master-node gradient information is divided into shards, where each of the shards runs independently.
    \end{enumerate*}
    Performing data-parallel distributed training asynchronously eliminates the need to synchronize the weight updates between the nodes; however, asynchronous training typically leads to a higher loss in the accuracy of the trained model. The authors of~\cite{qsparse} proposed Qsparse-local-SGD, which can be applied to synchronous and hybrid (combining both synchronous and asynchronous training) scenarios;
    in the latter case, nodes are divided into groups. The gradient transfer within a specific group is synchronous and asynchronous within the different groups.The Qsparse-local-SGD method achieved over 20$\times$ reduction in terms of the total number of bits transmitted in the training of Resnet50\cite{resnet} on the ImageNet\cite{imagenet} dataset, with an approximate $1\%$ loss in the final accuracy.
    
    \revised{Furthermore, a reduction in the total number of training iterations can also reduce the number of gradient transfers performed within the training and as a result, bring to the decrease of the total amount of transferred data. Reduction in the number of the iterations can be reached by an increase in the batch size. Given a fixed amount of memory in the graphical processing unit (GPU), an increase of the batch is possible if the space allocated for the model is decreased. Such a reduction is possible by means of implying model compression methods. A model can be compressed, employing quantization~\cite{wu2016quantized, jacob2018quantization, zhou2016dorefaquantize, rastegari2016xnorquantize, krishnamoorthi2018quantizing, zhou2017incrementalquantize}, pruning~\cite{pruning_3stage, pruning_old, gordon2018morphnetpruning, he2018amcpruning}, and knowledge distillation~\cite{hinton2015distilling, modeldistil}. Jacob et al.~\cite{jacob2018quantization} quantized the pre-trained FP32 network to a lower bit-depth precision using 8-bit integers for both weights and activations. Han et al.~\cite{pruning_3stage} propose to prune redundant connections using a three-step method. First, they learn important connections within a network, prune unimportant connections, and finally retrain the network. Authors of work~\cite{hinton2015distilling} introduced the knowledge distillation method, where knowledge from the larger model transferred to the smaller model. Typically, the downside of these methods is an inevitable degradation of performance. Moreover, model compression methods are usually applied on the already pretrained networks; hence most of them can not solve the problem of huge communication bandwidth needed for the distributed training.}

    Several approaches have been proposed to address the gradient communication problem, including gradient sparsification~\cite{Strom,DGC, qsparse}, quantization~\cite{Seide, QSGD, qsparse} and entropy coding~\cite{fekri_ndq, abdi2019reducing} of the gradient tensors; these approaches are proposed within the context of synchronous data-parallel distributed training. The core idea of gradient sparsification is to transfer a fraction of the gradient, depending on some importance metric. Deep Gradient Compression~\cite{DGC}, for example, follows the gradient sparsification approaches, achieving up to $99.9\%$ gradient sparsification without loss in accuracy. In the gradient quantization approaches, the gradients are being quantized before transferring. The authors of~\cite{QSGD}, for example, trained the ResNet-152~\cite{resnet} network to full accuracy on ImageNet~\cite{imagenet} 1.8$\times$ faster than the variant with full-presicion gradients. The other approach is to combine the aforementioned techniques, as it has been done in the method presented in~\cite{qsparse}. 
    
    Despite their efficiency, these approaches only explore the intra-node gradient redundancy by means of sparsification, quantization and entropy coding. In this work, we propose to exploit the correlation between the gradients of the distributed nodes in order to achieve further compression gains. It is worth mentioning that our approach can be combined with gradient sparsification and quantization. \revised{An attempt to explore the redundancies of gradients by different nodes was also made in~\cite{abdi2019reducing, chen2020scalecom}. The method described in~\cite{abdi2019reducing} is based on distributed source coding, realised by low-density parity check (LDPC) codes, which leads to an impractically high decoding latency and complexity. In contrast, our approach employs lightweight autoencoders for the compression of gradients, which substantially reduce the communication rate without compromising the encoding and decoding speed. In~\cite{chen2020scalecom} the authors also utilized the correlation between the gradient tensors. They introduced new Cyclic Local Top-k selection mechanism for the ring-allreduce communication pattern, where a set of indices for the gradient selection is the same for all of the nodes. Compared with this approach, our LGC framework is able to provide higher compression ratios because of the further utilization of the correlation and introduction of autoencoder-based distributed compressors.}
    In summary, the contributions of this work are:
    \begin{itemize}
    
    \item we study experimentally the statistical dependency among gradients of distributed nodes using information theoretic metrics and show that there is a considerable rate reduction that can be achieved if these correlations are exploited. 
    
    \item we propose a novel framework for performing distributed compression of the gradients, coined Learned Gradient Compression (LGC). Our framework uses lightweight autoencoder models to compress gradients by leveraging the correlation between them. We propose different instances of our framework for the \textit{parameter server} and \textit{ring-allreduce} distributed communication patterns. To the best of our knowledge, this is the first attempt to use autoencoders, which capture the correlation across distributed signals, for compressing gradients.  

    \item we experimentally evaluate our method against different benchmarks--including uncompressed gradients (baseline method), and the state-of-the-art DGC~\cite{DGC} and ScaleCom~\cite{chen2020scalecom} methods---and show that we systematically achieve significant rate reductions for different tasks, models and datasets. 
\end{itemize}

The remainder of the paper reads as follows:~\secref{sec:relatedWork} presents different protocols for distributed training and discusses the related work.~\secref{sec:InfoPlane} analyses the correlation of gradient tensors in data-parallel distributed training through the lens of information theory. The result of the analysis motivates the use of the proposed gradient compression autoencoders, presented in~\secref{sec:CompressNetworkArchitecture}.~\secref{sec:TheFramework} describes how the autoencoder models are used  within the proposed LGC framework. We present our experimental results in~\secref{sec:Experiments} and conclude our work in~\secref{sec:Conclusion}.

\begin{figure}[t]
    \centering
    \includegraphics[width=0.65\linewidth]{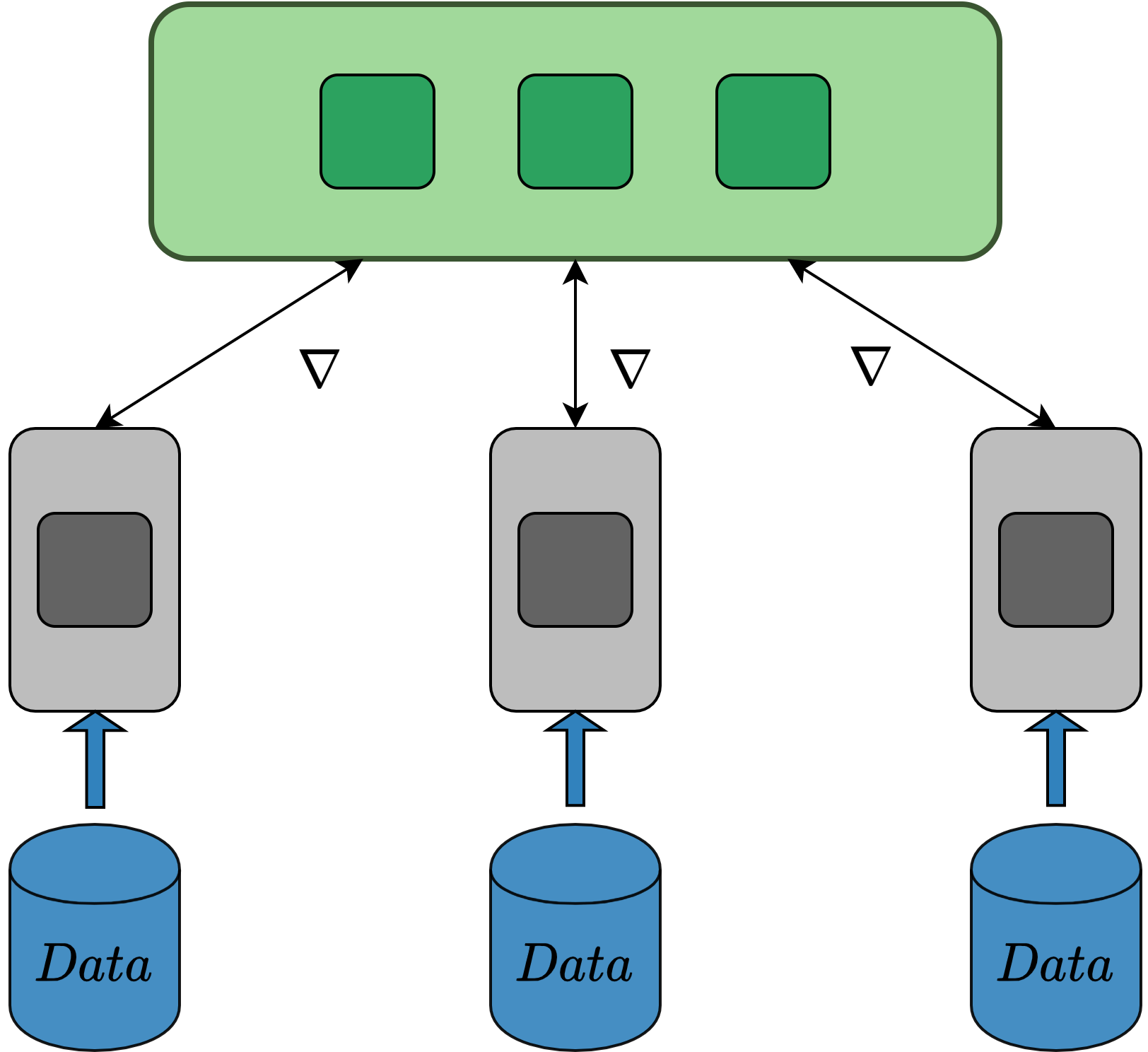}
    \caption{Illustration of the parameter server communication pattern. $\nabla$ denotes a locally computed gradient tensor.}
    \label{fig:PS_description}
\end{figure}

\begin{figure}[t]
    \centering
    \includegraphics[width=0.6\linewidth]{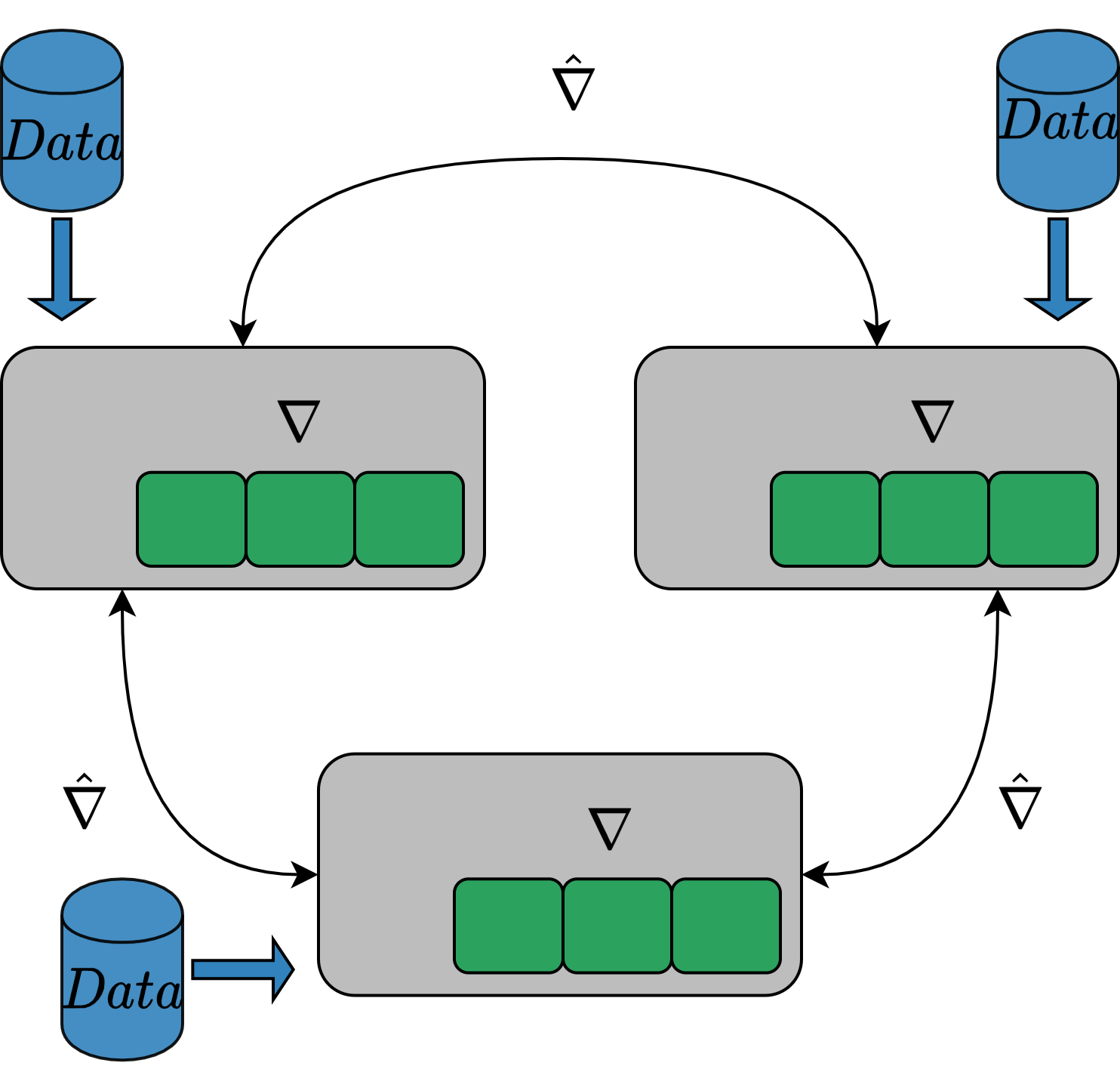}
    \caption{Illustration of the ring-allreduce communication pattern. $\nabla$ is a locally computed gradient tensor and $\hat{\nabla}$ is a part of the gradient tensor, which is exchanged between the neighbors.} 
        \label{fig:RAR_description}

\end{figure}

%%%%%%%%%%%%%%%%%%%%%%
\section{Setup and Related Work}
\label{sec:relatedWork}
%%%%%%%%%%%%%%%%%%%%%%
    \subsection{Setup}
        For the data-parallel distributed training, we consider two of the most well-established protocols of distributed communication, namely, the \textit{parameter server}~\cite{parameter-server} and the \textit{ring-allreduce}~\cite{ring-Allreduce,reduce_optimal} protocols. In the \textit{parameter server} scenario (see its schema in \figref{fig:PS_description}), the nodes are divided into two types: the worker nodes, which contain a replica of the neural network to be trained and calculate the gradient tensor of it using the available data, and the master node\footnote{Depending on the application scenario, one of the worker nodes can take over the responsibility of the master node. This eliminates the required resource allocation for a standalone master node in the system.}, which receives the gradient tensors from the worker nodes, performs a reduction operation and sends back the updated gradient tensor to the worker nodes.  In the \textit{ring-allreduce} protocol (see its schema in \figref{fig:RAR_description}), there is no master node and the calculation of the global gradient tensor is performed through the exchange of the local gradient tensors between neighboring nodes. In each node, the gradient tensor is divided into $K$ parts, where $K$ is the number of nodes. In the first phase of the communication, each node sends and receives a part of those gradient tensors. The received values are added to the corresponding values already available at the node. The transfer operation lasts for $K-1$ iterations. These $K-1$ information exchanges are  organized in a way that after each of them each node holds a part of the final gradient tensor, e.g., the gradient values at the same indices across all nodes. Subsequently, for another $K-1$ iterations, the nodes exchange those parts of the final gradient tensor among them. In the \textit{ring-allreduce} protocol, each of the $K$ nodes communicates with two of its neighbors $2\times(K-1)$ times. The bottleneck in both approaches is the size of the gradient tensor, which is huge in the case of very deep models with a lot of parameters; hence, these gradient transfers can significantly decelerate the training.
   
    \subsection{Related Work}
 
    Various solutions have been proposed to address the problem of the huge communication bandwidth required for distributed training. Bellow, we categorize the solutions into a few major themes.\\
    
    \noindent\textbf{Sparsification:} The study in~\cite{sparsification_Fikri} proposed to sparsify the gradient tensor by replacing values below a certain threshold with zeros. The method resulted in a $99\%$ gradient sparsification during the training of a fully-connected deep neural network on MNIST~\cite{mnist}, achieving an accuracy of $99.42\%$. Alternatively, Sparse Gradient Descent (Sparse GD)~\cite{Strom} applied top-$k$ gradient selection to obtain sparse gradients. 
    The study in~\cite{wangni2018gradient} proposed a sparsification method that randomly drops out coordinates of the stochastic gradient vectors and amplifies the remaining coordinates to ensure that the sparsified gradient is unbiased.\\
    
    \noindent\textbf{Quantization:} The authors of~\cite{QSGD} proposed a family of algorithms for lossy gradient compression based on quantization, called QSGD. They trained the ResNet-152~\cite{resnet}  network  to full accuracy on ImageNet~\cite{imagenet}, 1.8$\times$ faster than the variant with full-precision gradients. Dithered quantization followed by adaptive arithmetic encoding of the quantized gradients was proposed in~\cite{fekri_ndq}. When training AlexNet~\cite{cnn1} on Cifar10~\cite{cifar}, the method reduced the communication bits per node from $8531.5$ bits to $422.8$ bits, achieving $65.6\%$ accuracy (which is lower by $2.6\%$ compared with the non-distributed version of the network). In the context of federated learning,~\cite{google_federated}, \cite{federated_felix} reported a combination of top-$k$ gradient sparsification in the downstream communication (i.e., from the master to the workers) with ternary gradient quantization in the upstream communication, achieving $85.46\%$ accuracy on Cifar10 using a modified version of VGG~\cite{vgg} (coined VGG11). \revised{Furthermore, Amiri \etal~\cite{quant_gmfederated} proposed to quantized the updates being sent from the central node to all of the devices by exploiting the knowledge of the last global model estimate available at the devices as side information. }
    \\

    \noindent\textbf{Error correction:} Error correction techniques, which compensate for the errors introduced by the compression of gradients, have also been proposed.
    The study in~\cite{ef-sgd} introduced EF-SGD, an error-feedback (EF) mechanism applied to the training with SGD (on a single graphics processing units (GPU)). The idea was to combine residuals of the compression from the previous iteration with the current gradient before performing an update of the parameters.
    The authors of~\cite{blockwise-sgd} applied the EF mechanism to data-parallel distributed training. The so-called Dist-EF-SGD method quantizes the gradient of each layer to 1 bit and combines it with the residual of the compressed gradient of the previous iteration. That way it achieves a 32$\times$ reduction of the communication cost, while retaining the same test accuracy on ResNet50~\cite{resnet} trained on the ImageNet \cite{imagenet} dataset, and reducing the total time by $46\%$ compared to the case that uncompressed weights are used during training.  Alternatively, the study in~\cite{Seide} proposed an EF mechanism where the gradients are quantized to 1 bit and the quantization errors are added to the gradients of the next batch.
    The combination of gradient compression via sparsification with error correction has been studied in~\cite{DGC, stich2018sparsified}. The study in~\cite{stich2018sparsified} proposed MEM-SGD, a technique that keeps track of the accumulated errors due to top-$k$ gradient sparsification and adds them back to the gradient estimator before each transmission. MEM-SGD converges at the same rate as SGD on the convex problems, whilst reducing the communication overhead by a factor equal to the problem dimensionality.
    The authors of~\cite{DGC} proposed Deep Gradient Compression (DGC), in which only important gradients (\ie based on the magnitude of the gradient) are sent at each node at every iteration. The rest of the gradients are accumulated at each node using momentum correlation, a sort of error correction mechanism, and sent when they pass the threshold of importance. DGC achieves up to $99.9\%$ sparsification of gradients without loss of accuracy when training ResNet101~\cite{resnet} on Cifar10~\cite{cifar}.
    \revised{In the field of federated learning, Athe authors of~\cite{fetchsgd} proposed a method called FetchSGD, where they moved momentum and error accumulation from clients to the central aggregator. This transition of the accumulation operation was possible because of the use of so-called Count Sketch randomized data structure. Advantage of the Count Sketch is that it can compress a vector by randomly projecting it several times to lower-dimensional spaces, such that high-magnitude elements can later be approximately recovered. Further, in~\cite{philippenko2020artemis}, the authors proposed an Artemis framework in which, using an error correction mechanism, they compress both upstream and downstream communication.} \\
    
   \revised{\noindent\textbf{Gradients similarity:} Few works in the field of distributed training concentrate on the study of correlation of the gradients within the different nodes.
    Authors of~\cite{chen2020scalecom} proposed the Scalable Sparsified Gradient Compression (ScaleCom) method, within which they explored the similarity between the gradient residuals within a framework of distributed training performed combined with the error-correction techniques. They had explored the cosine distance between the gradient residuals at the different nodes. They showed that the distance decreases fast over the iterations, i.e., local memory similarity is improved quickly and stays correlated over much of the training process. Based on this similarity, the author of the 5 proposed new Cyclic Local Top-k (CLT-k) compressor for all-reduce distributed training. The proposed method works as follow: one of the workers sorts its error-feedback gradient, obtains its local top-k indices and further, all other workers follow the leading worker’s top-k index selection for compressing their local error-feedback gradients. Worth to mention that similar selection process is also used in our LGC framework for the ring-allreduce communication pattern. Moreover, because of the proposed autoencoder, our framework can provide higher compression rate compared with the ScaleCom.
     Leveraging the similarities in the gradient information transmitted by different workers has also been proposed by Abdi and Fekri~\cite{abdi2019reducing}. Their compression method follows the distributed source coding~\cite{slepian1973noiseless,wyner1976rate,berger1996ceo} principles: The gradients of the different nodes are modeled as noisy versions of the true gradient, where the noise is assumed~\textit{i.i.d.}~Gaussian (this refers to the CEO problem~\cite{berger1996ceo} in information theory). The technique applies asymmetric Wyner-Ziv coding~\cite{wyner1976rate}, where the gradients from a group of nodes are intra-encoded and used as side information to decode the gradients from the rest of the nodes. For the latter group of nodes, compression is performed using nested scalar quantization followed by syndrome encoding~\cite{pradhan2003distributed} realized by low-density parity check (LDPC) codes.  However, applying LDPC decoding of the gradients per training iteration induces a significant decoding complexity and latency, limiting the practical application of the method.}

    In this work, we introduce a new compression framework that uses autoencoders to capture the correlation across gradients of distributed nodes. Unlike the method in~\cite{abdi2019reducing}, our approach (\textit{i}) assumes that gradients share a common information component and differ by an innovation component; (\textit{ii}) uses lightweight autoencoders that induce limited computational complexity at the encoder and decoder side.

    To the best of our knowledge, our work is the first to propose the use of autoencoder models for the compression of gradients in distributed training, which also capture the gradient correlations across nodes. Autoencoder models have been proposed for image compression~\cite{drasic, sourcecoding_nn} and image compressed sensing~\cite{DCS}. These models, however, use deep CNN and recurrent neural network (RNN) architectures for compression (as opposed to our lightweight models), which are not favorable in the distributed training setting where fast encoding and decoding is paramount. Furthermore, while the autoencoder architecture in~\cite{drasic} also considers the correlation across images from distributed cameras, the model architecture, the data at hand (images in~\cite{drasic} versus gradients in our case) and the setting (distributed camera communication in~\cite{drasic} versus distributed training in our work) are different than ours.

    %%%%%%%%%%%%%%%%%%%%%%
    \section{Information Plane of Gradients within a Distributed Training}
    \label{sec:InfoPlane}
    %%%%%%%%%%%%%%%%%%%%%%
    \revised{In this section, we demonstrate the importance of leveraging the correlation (a.k.a., redundancy) of gradients in distributed nodes in the reduction of the compression rate from an information-theoretic point of view. To this end,} we analyze the statistical dependencies among the gradient tensors produced by different computing nodes within distributed training using information-theoretic measures; namely, the marginal and conditional entropy, and the mutual information (MI).
    The calculation of these measures relies on the estimation of the underlying probability density function (pdf) of the observed variables. Different approaches can be used to calculate the marginal and conditional pdf of gradient tensors including histogram, parametric, and non-parametric estimation. In this study, we consider the basic method of calculating the histogram as we are merely interested in corroborating that gradients across distributed nodes are correlated.   
    
    We conduct experiments using deep neural networks trained on two image transformation tasks, image classification, and semantic segmentation. For the image classification task, we train the Resnet50~\cite{resnet} model on the Cifar10~\cite{cifar} dataset and, for the semantic segmentation task, we train the PSPNet~\cite{pspnet} model on the CamVid~\cite{camvid} dataset. In both cases, training is performed on two distributed nodes using synchronous SGD.
    
    Let $g_{l,1}^{(i)}$ and $g_{l,2}^{(i)}$ denote gradient vectors (vectorized tensors) of distributed nodes, where $l=1, 2, \dots, L,$ indexes the convolutional layer in the model, $1$ or $2$ identifies the computing node that calculates the gradient, and $i=0, 1, \dots, N,$ denotes the training iteration. Each gradient is discretized using the uniform quantizer with $2^{32}$-level (32 bit) accuracy. The quantized gradients are then used to approximate the marginal and the joint densities, respectively, $p_{g_{l,2}}^{(i)}$ and $p_{g_{l,1}, g_{l,2}}^{(i)}$, using the histogram method. Using the estimated densities, we estimate the marginal and conditional entropy, denoted by $H(g_{l,2}^{(i)})$ and $H(g_{l,2}^{(i)}|g_{l,1}^{(i)})$, respectively. The MI between the two gradient vectors, at iteration $i$, is then calculated as 
    \begin{align}
        I(g_{l,1}^{(i)}; & g_{l,2}^{(i)}) 
            \nonumber 
            \\ 
            & = H(g_{l,2}^{(i)}) - H(g_{l,2}^{(i)}|g_{l,1}^{(i)}) 
           \nonumber 
            \\ 
            & = -\sum p_{g_{l,2}}^{(i)} \log_{2}p_{g_{l,2}}^{(i)} 
            + \sum p_{g_{l,1}, g_{l,2}}^{(i)} \log_2 \frac{p_{g_{l,1}, g_{l,2}}^{(i)}}{p_{g_{l,1}}^{(i)}}
        \end{align}

\begin{figure}[t]   
    \centering
    \subfloat[]{\includegraphics[width=0.4\textwidth]{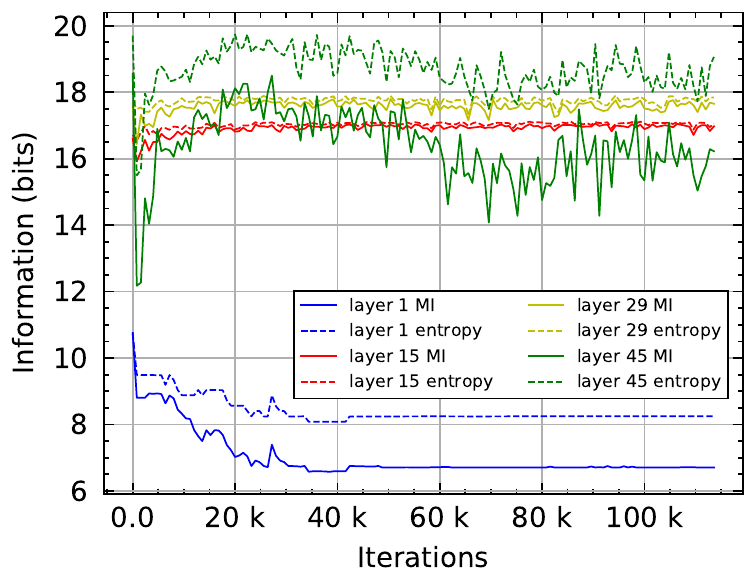}
    \label{fig:mu_resnet50}}
    \hfill
    \subfloat[]{\includegraphics[width=0.4\textwidth]{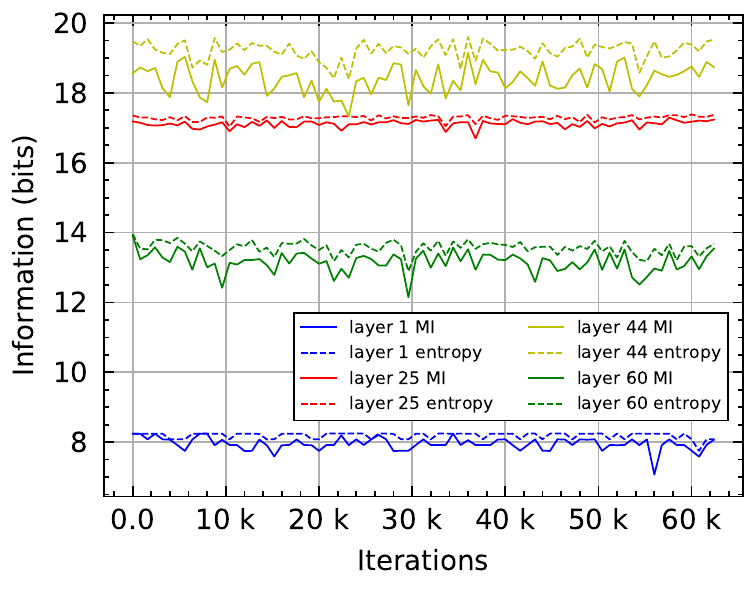}
    \label{fig:mu_pspnet}}
    \caption{The mutual information (solid lines) and the marginal entropy (dotted lines) between gradient tensors of the same layer on the different nodes, through the training iterations, for distributed training of (a) the Resnet50 and (b) the PSPNet models.}
    \label{fig:mu_pspnet_resnet}
    \end{figure}

    \Figref{fig:mu_pspnet_resnet} depicts the marginal entropy $H(g_{l,2}^{(i)})$ and the MI $I(g_{l,1}^{(i)}; g_{l,2}^{(i)})$ for different pairs of layers in the models throughout the training iterations. We observe that the MI values are high; specifically, approximately $80\%$ of the average information content (i.e., the entropy) contained in the layer's gradient tensor at every iteration is common for both nodes. This means that there is a significant amount of information that can be obtained from one gradient tensor about the other per iteration. 
    Note that we observe similar trends between the marginal entropy and the MI across the iterations,
    which indicates that this correlation can be leveraged independent of the changes in the information content of the gradient tensor.
    
\begin{figure}[t]   
        \centering
            \subfloat[]{\includegraphics[width=0.4\textwidth]{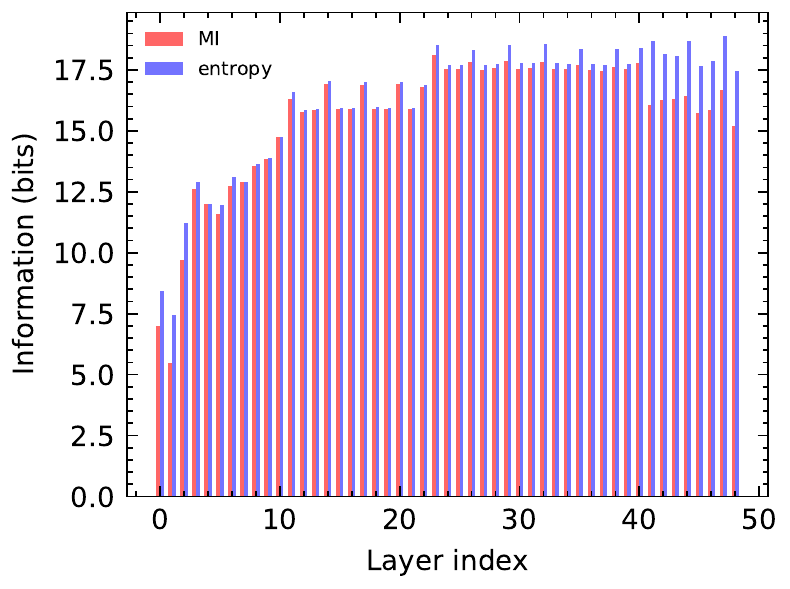}
            \label{fig:mu_hist_resnet50}}
            \hfill
            \subfloat[]{\includegraphics[width=0.4\textwidth]{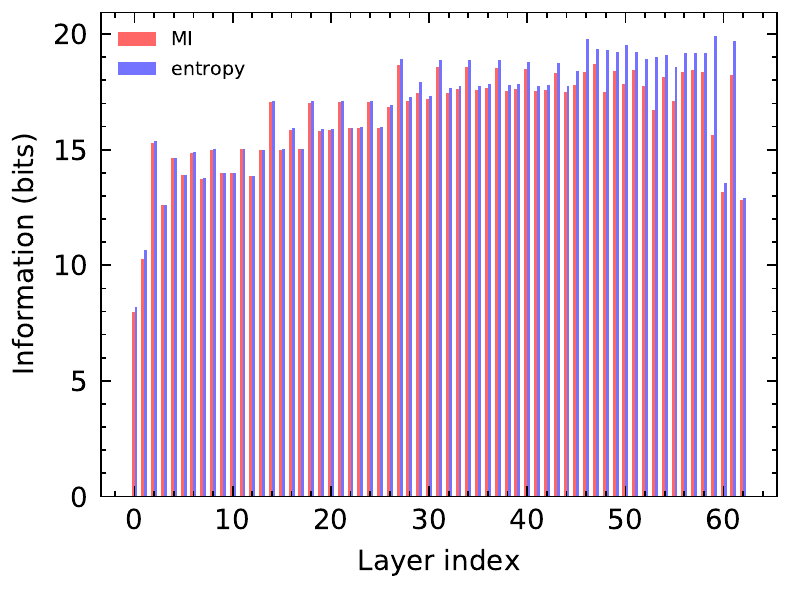}
            \label{fig:mu_histogram_pspnet}}
        \caption{The mean marginal entropy and the mean MI per layer, averaged over the iterations, for (a) the Resnet50 trained on Cifar10 and (b) the PSPNet trained on CamVid.}
        \label{fig:hists}
    \end{figure}

    \Figref{fig:hists} illustrates the mean marginal entropy $\Big( \frac{1}{N}\sum_{i}^{N}H(g_{l,2}^{(i)}) \Big)$ and the mean MI $\Big( \frac{1}{N}\sum_{i}^{N} I(g_{l,1}^{(i)}; g_{l,2}^{(i)}) \Big)$ for all layers $l=1, 2, \dots, L$, averaged over the iterations. One can observe that the mean marginal entropy increases with the layer index, which is due to the increase in the number of parameters in corresponding layers. 
    Interestingly, the mean MI is systematically comparable with the mean marginal entropy for almost every layer. The lowest amounts of MI between the gradients of the two nodes--namely, the higher discrepancy with the entropy--were obtained in the first and the last layers. This is because the gradients of these layers are influenced the most by the input training samples and the ground truth labels in the mini-batch. We can also notice some systematic peaks in the values of the entropy and MI. These peaks can be explained based on the architecture of each model. Both models contain residual connections, where information between layers is combined through the element-wise summation operation. 
    Thus, the amount of information should increase after the addition operation. This could be also confirmed from the results. Specifically, we observe that the layers with a higher amount of information are those that come after residual connections.

    \revised{Our empirical results suggest that the gradient information per layer at each node consists of two parts, the \textit{common information} shared across all gradient vectors and the \textit{innovation information} that is specific for each node. Moreover, the \textit{common information} holds a substantial amount of the gradients' information. In other words, in distributed training, there is a significant amount of redundant information --currently being sent at each iteration from all the nodes--which if eliminated can further reduce the communication rates without affecting the performance of the trained model. For the simplicity, within the proposed framework, we assume that innovation component can be captured by performing top-k selection (with the sparsity rate of $0.001\%$) over the gradient tensors of each node.}
    
    In what follows, we present two instances of our LGC framework that leverage this information in the \textit{parameter server} and the \textit{ring-allreduce} communication pattern.

    %%%%%%%%%%%%%%%%%%%%%%
    \section{Autoencoders for Gradient Compression}
    \label{sec:CompressNetworkArchitecture}
    %%%%%%%%%%%%%%%%%%%%%%

      In this section, we introduce two autoencoder architectures for performing lossy (\ie a reconstructed signal differs from the original one) distributed compression of gradients. 
      We design our architectures based on our empirical observations in~\secref{sec:InfoPlane}. Specifically, our models aim at capturing the fact that gradients at the same layers of the different nodes are highly correlated. 
      Two such autoencoder models are described, one for the~{parameter server} (see~\secref{sec:ParameterAutoencoder}) communication patterns and the other for the~{ring-allreduce} (see~\secref{sec:RingAutoencoder}).
     
      \subsection{Autoencoder Based on Decoupling of the Gradients for the Parameter Server Communication Pattern }
      \label{sec:ParameterAutoencoder}
      
      Consider a set of gradient tensors, each unfolded in the form of a vector $g_{k}$, $k=1, ..., K$. These gradients correspond to the same training iteration of synchronous SGD across $K$ nodes, where the index of the iteration is omitted for simplicity. According to our modelling approach, each of the gradient vectors can be expressed as\giannis{:}
      \begin{equation}
          g_{k} = g^{cp}_{k} + g^{I}_{k},
      \end{equation}
      where $g^{cp}_{k}$ is the common part that is shared by the gradient vectors and $g^{I}_{k}$ is the innovation component that is unique to each gradient vector. 
      The proposed autoencoder model for the compression and the reconstruction of the gradients consists of one encoder, $E_{c}$, and $K$ decoders, $D_{c}^{k}$, $k=1, ..., K$.
      The encoder encodes one of the gradients~$g_{i}$, with $i \in (1, ..., K)$, into a \textit{compressed common representation}:
      \begin{equation}
        g^{c} = E_{c}(g_{i}). 
      \end{equation}
      Each of the $K$ decoders inputs the compressed common representation, $g^{c}$, and combines it with the innovation information of the $k$-th gradient, $g^{I}_k$, to obtain the corresponding reconstructed gradient vector,
      \begin{equation}
        g^{rec}_{k} = D_{c}^{k}(g^{c}, g^{I}_k).
      \end{equation}
      During the training process of the autoencoder, in order to obtain the compressed common representation $g^{c}$ from the gradient vectors, all gradients are fed to the encoder $E_{c}$. The following similarity loss (\ie the Euclidean distance between the compressed representations of the gradients) is minimized:
      \begin{equation}
        L_{sim} = \sum_{k=0}^{K} \sum_{m=0 (m\neq k)}^{K} || (E_{c}(g_{k}) - E_{c}(g_{m}))||^2_2 \quad 
      \end{equation}
      In order to reconstruct the original gradients from the compressed ones, the following reconstruction loss is also applied to the output of the decoders $D_{c}^{k}$:
      \begin{equation}
        L_{rec} = \sum_{k=0}^{K} || (g_{k} - g^{rec}_k)||^2_2 \quad 
      \end{equation}
      The final loss function therefore consists of two terms, the reconstruction loss and the similarity loss, that is,
      \begin{equation}
        L = \lambda_{1} L_{rec} + \lambda_{2}L_{sim}.
        \label{eq:loss_ps}
      \end{equation}
      Within the data-parallel distributed training, the  training of the autoencoder (see~\figref{fig:autoencoder}(a)) is performed at the master node as follows. The master node receives the gradient vectors from the worker nodes and passes them sequentially to the encoder $E_{c}$. The encoded representations are used to calculate the similarity loss $L_{sim}$. Furthermore, one of the encoded representations (chosen randomly at each iteration) is combined with the innovation components, $g^{I}_{k}$, $k=1,\dots,K$, at the corresponding decoders $D_{c}^{k}$ to reconstruct the gradients and compute the $L_{rec}$ loss~(see~\figref{fig:ps_train_autoencoder}).
      
      \subsection{Autoencoder Based on Aggregation of the Gradients for the Ring-Allreduce Communication Pattern}
      \label{sec:RingAutoencoder}
      The final goal of one iteration of distributed training is to obtain an aggregated gradient from the gradients of distributed nodes and send it back to all nodes so that the parameters are further updated. 
      To this end, we design an architecture where, as a first step, a separate compressed representation for each of the gradients is obtained and then, in a next step, the compressed representations are averaged. 
      One aggregated gradient vector, which approximates the average of the gradients of distributed nodes, is then reconstructed from the averaged compressed representations. 
      In this case, the autoencoder consists of one encoder $E_{c}$, which is responsible for the compression of the gradient vectors, and one decoder $D_{c}$ %, with the help of which the final aggregated gradient is produced.  
      that is responsible for the final aggregated gradient vector.
      For the group of the gradients, $g_{k}$, $k=1, ..., K$, the compressed representations, $g^{c}_{k}$, obtained by the encoder $E_{c}$ are given using:
      \begin{equation}
          g^{c}_{k} = E_{c}(g_{k}).
      \end{equation}
      The intermediate representation $g^{avg}$, which is obtained as\giannis{:}
      \begin{equation}
          g^{avg} = \frac{1}{K} \sum_{k=1}^K E_{c}(g_{k}),
      \end{equation}
      is fed to the decoder $D_{c}$, which reconstructs the final aggregated gradient vector as:
      \begin{equation}
          g^{rec} = D_{c}(g^{avg}).
      \end{equation}
      In the training process, the following reconstruction loss is applied on the final aggregated gradient to minimize the distance between the output of the decoder and the average of the gradient vectors:
      \begin{equation}
      L_{rec} =  || (g^{rec} - \frac{1}{K} \sum_{k=1}^K g_{k})||^2_2.
      \end{equation}

\begin{figure*}[ht]
    \centering
    \includegraphics[width=0.7\textwidth]{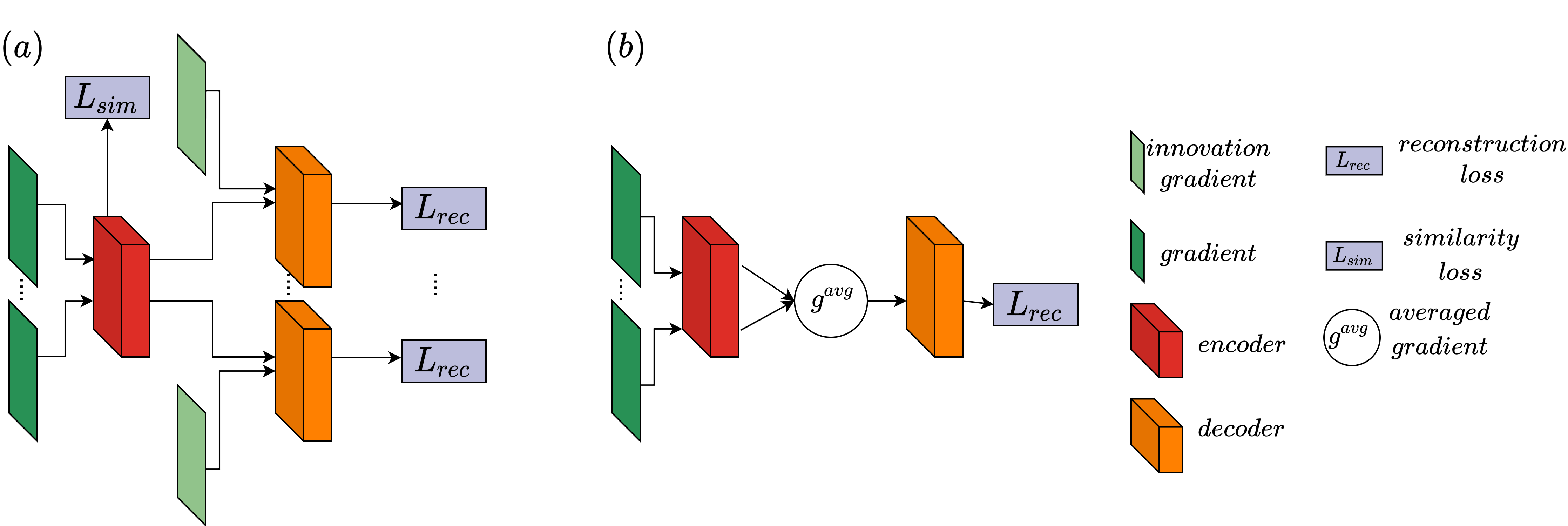}
    \caption{The architecture of the proposed autoencoders for the two different distributedcnode communication scenarios: (a) in the parameter server scenario, gradients are fed sequentially to the encoder and, at each decoder, the innovation gradients are concatenated with the intermediate output before the last convolutional layer; (b) in the ring-allreduce scenario, the vectors are fed to the encoder sequentially, and after that, the outputs of the encoder form the averaged gradient, which is passed to the decoder for the reconstruction. Note that the number of gradients, decoders and the reconstruction losses needs to be equal to the number of distributed nodes}
    \label{fig:autoencoder}
    \end{figure*}

    Within the data-parallel distributed training, the  training of the autoencoder (see~\figref{fig:autoencoder}(b)) is performed as follows. One node obtains the gradient vectors, $g_{k}$, and sequentially feeds them into the encoder $E_{c}$, which in turn constructs the intermediate representation, $g^{avg}$, and passes this representation to the decoder $D_{c}$. The decoder reconstructs the gradient $g^{rec}$, which is used to calculate the loss $L_{rec}$ (see~\figref{fig:rar_train}).
    
    \subsection{Distributed Autoencoders Architecture}
        In our distributed compression approach, the proposed autoencoder networks (see \secref{sec:ParameterAutoencoder} and \secref{sec:RingAutoencoder}) consist of convolutional and deconvolutional layers. The kernels of the convolutional and deconvolutional layers are one-dimensional (1D) since the inputs of the networks are vectors.
        This approach reduces the number of network parameters compared with the conventional 2D kernels by approximately $60\%$.
        
        Regarding the autoencoder architecture that is based on decoupling of the gradients (see ~\secref{sec:ParameterAutoencoder}), the
        encoder, $E_c$, which takes as input the vector $g_{k}$, comprises 5 convolutional layers with 1D kernels and a non-linearity in the form of the leaky-relu~\cite{leaky-relu}. The hyperparameters of the encoder are reported in~\tabref{table:encoder_params}. The decoder, $D_c$, takes as input the compressed representation produced by the encoder and performs an upsampling operation using 5 deconvolutional layers. Before the final convolutional layer, the intermediate representation is concatenated with the innovation component, $g^{I}_{k}$, which consists of the top-magnitude values of $g_{k}$ and zeros elsewhere; more details on how the innovation component is constructed are provided in~\secref{sec:TheFramework}. The hyperparameters of the decoder are reported in~\tabref{table:decoder_params}.

        Regarding the autoencoder that is based on the aggregation of the gradients (see~\secref{sec:RingAutoencoder}), the architecture of the encoder, $E_c$, and the decoder, $D_c$, are the same with the only difference that there is no concatenation operation and the input of the decoder is the average, $g^{avg}$, of the compressed gradients.
        
        \begin{table}[t]
            \begin{center}
            \caption{Parameters of the convolutional layers of the encoder $E_c$.}
                \begin{tabular}{c||c|c|c}
                \thickhline
                Layer &  \# Filters & Kernel Size & Stride \\ 
                \thickhline
                conv1  & 64 & $1\times 3$ & $1\times 2$ \\
                \hline
                conv2  & 128 & $1\times 3$ & $1\times 2$ \\
                \hline
                conv3  & 256 & $1\times 3$ & $1\times 2$ \\
                \hline
                conv4  & 64 & $1\times 3$ & $1\times 2$ \\
                \hline
                conv5  & 4 & $1\times 1$ & $1 \times 1$ \\
                 \thickhline
                \end{tabular} 
            \label{table:encoder_params}
            \end{center}
        \end{table}

        \begin{table}[t]
            \begin{center}
            \caption{Parameters of the deconvolutional and convolutional layers of the decoder $D_c$.}
                \begin{tabular}{c||c|c|c}
                \thickhline
                Layer & \# Filters & Kernel Size & Stride \\ 
                \thickhline
                deconv1  & 4 & $1\times 3$ & $1\times 2$ \\
                \hline
                deconv2  & 32 & $1\times 3$ & $1\times 2$ \\
                \hline
                deconv3  & 64 & $1\times 3$ & $1\times 2$ \\
                \hline
                deconv4  & 128 & $1\times 3$ & $1\times 2$ \\
                \hline
                deconv5  & 32 & $1\times 3$ & $1\times 2$ \\
                \hline
                conv  & 1 & $1\times 1$ & $1 \times 1$ \\
                 \thickhline
                \end{tabular} 
                \label{table:decoder_params}
            \end{center}
        \end{table}

    %%%%%%%%%%%%%%%%%%%%%%
    \section{Distributed Learning Framework}
    \label{sec:TheFramework}
    %%%%%%%%%%%%%%%%%%%%%%
    
    Assume a system with multiple GPUs, consisting of $K$ processing nodes. The goal is to train a model (with layers $l=1, \dots,L$) in a data-parallel distributed manner (where replicas of the model are located at each node) using synchronous SGD. Without loss of generality, it can be considered that the model is a fully convolutional neural network~\cite{cnn2,cnn1}. 
    At each training iteration, a gradient tensor $\nabla_{k,l}$ is produced for each layer $l=1,\dots, L$ of the neural network and for each node $k=1,\dots, K$. Each such gradient tensor is unfolded in the form of a vector $g_{k,l} \in \mathbb{R}^{n_{l}}$, where $n_{l} = k^{h}_{l} \cdot k^{w}_{l} \cdot f_{l-1} \cdot f_l$, with $k^h_{l}, k^{w}_{l}$ denoting respectively the kernel height and the kernel width at the layer \textit{l}, $f_{l-1}$ the number of filters in the previous layer and $f_{l}$ the number of filters in the current layer.

    %%%%%%%%%%%%%%%%%%%%%%
    \subsection{Gradient Sparsification Process}
     \label{sec:GradientSparsification}
    %%%%%%%%%%%%%%%%%%%%%%
    Within the proposed LGC framework, in order to reduce the amount of gradient information sent from each node, a certain amount of values are selected from each vector, $g_{k, l}$, in two different ways depending on the communication pattern. Specifically, under the \textit{parameter server} communication pattern, the framework extracts the $\alpha\%$ of the values 
    in $g_{k, l}$ with the highest magnitude and constructs the vector $\tilde{g}_{k,l} \in R^{\mu_{l}}$, where $\mu_{l} = \frac{\alpha}{100}\cdot n_{l}$. The top-magnitude gradient selection process is repeated for all layers and concatenates the $\tilde{g}_{k,l}$, $l=1,\dots,L$, vectors together to form the vector $\tilde{g}_{k} \in R^{\mu}$, with $\mu = \sum_{l=0}^L\mu_{l}$. This process is performed independently at each node $k=1,\dots,K$ with $\alpha$ fixed across the nodes (typically $\alpha=0.1$). 
    Alternatively, under the \textit{ring-allreduce} communication pattern, our framework selects a node $k$ randomly for each iteration. This selected node in turn extracts the $\alpha\%$ of the values 
    in $g_{k, l}$ with the highest magnitude and constructs the vector $\tilde{g}_{k,l} \in R^{\mu_{l}}$. As in the previous setting, a vector $\tilde{g}_{k} \in R^{\mu}$ is created by concatenation of the top-magnitude gradients of all layers. The selected node then shares the indices of the extracted gradient values to all remaining nodes in the network, which in turn construct the corresponding vectors $\tilde{g}_{k'} \in R^{\mu}$, where $k'=1,\dots,K$ and $k'\ne k$. 
    
    Hence, while in the \textit{parameter server} communication pattern, each node is free to select independently the important gradients, in the \textit{ring-allreduce} pattern all nodes select gradients in the same positions as indicated by the selected node for each iteration. The transferred indices are entropy encoded--using the DEFLATE compression method~\cite{deflate_pattent}--and their rate is taken into account in the total rate calculation (see Section~\ref{sec:Experiments}).
    
    In both patterns (\ie parameter server and ring-allreduce), the selected gradients $\tilde{g}_{k}$ are passed to an encoder, the architecture of which is described in Section~\ref{sec:CompressNetworkArchitecture}, which performs further compression. Finally, the remaining non-selected gradients are being accumulated using a momentum correlation similar to the method described in~\cite{DGC}.
  
    \begin{figure}[t]
        
        \centering
        \includegraphics[width=0.8\linewidth]{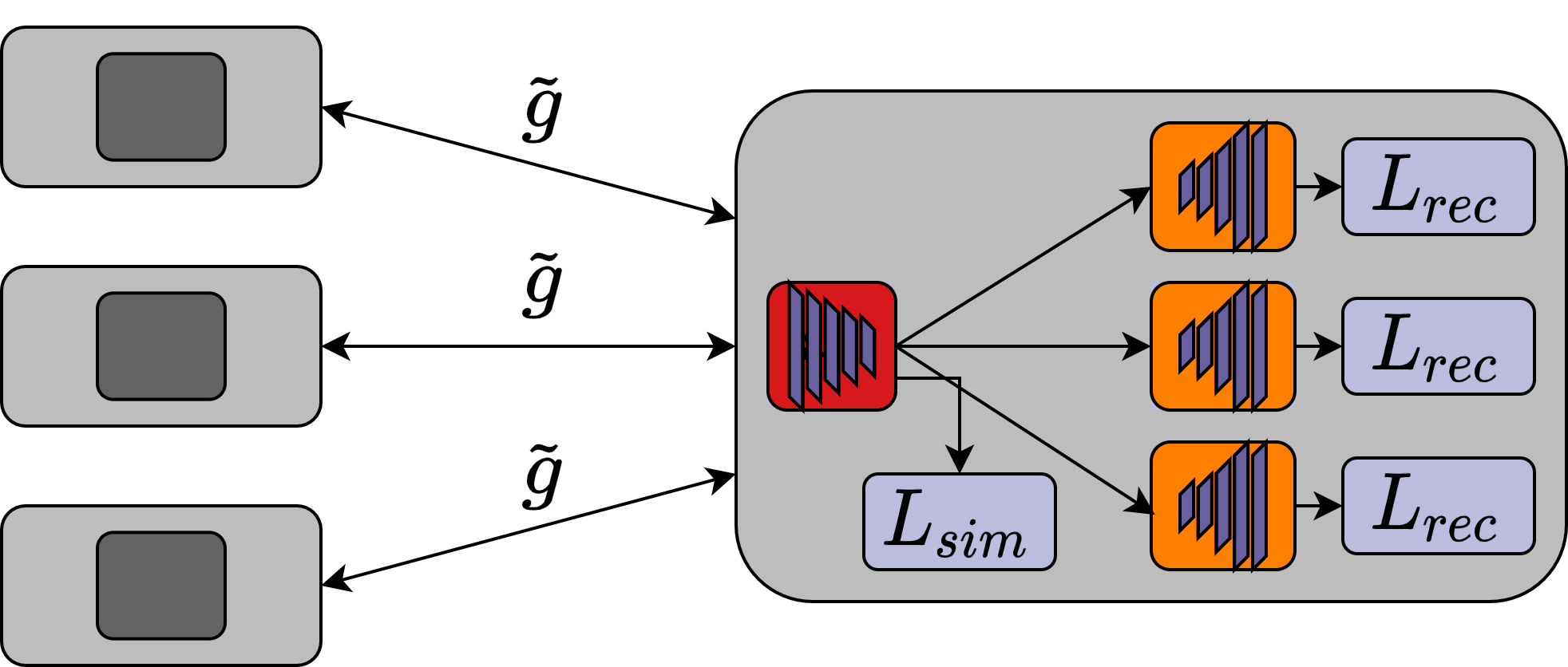}  
        \caption{Training of the proposed autoencoder for the parameter server communication pattern. $\tilde{g}$ is the gradient vector that is constructed using the top-magnitude gradient values. Note that in the training phase of the autoencoder, the innovation gradient vector $\tilde{g}^{I}$ is extracted on the master node from the $\tilde{g}$ gradient vector.
        }
    \label{fig:ps_train_autoencoder}
    \end{figure}

    \begin{figure}[t]
        \centering
        \includegraphics[width=0.6\linewidth]{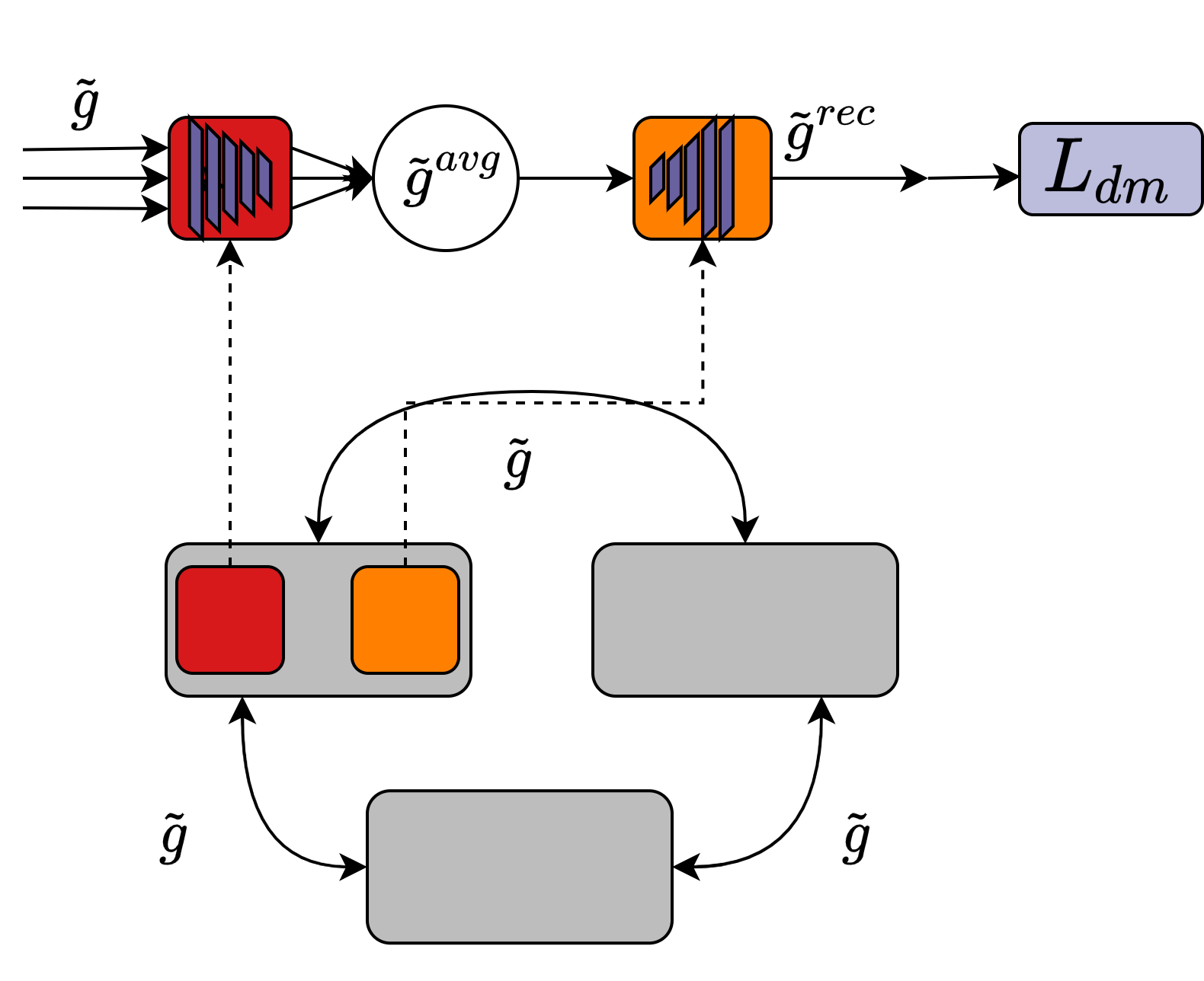} 
        \caption{Training of the proposed autoencoder for ring-allreduce communication pattern. Note that our framework selects randomly at each iteration the node that is responsible for extracting the values with the highest magnitude in order to construct the $\tilde{g}$ gradient vector.}
    \label{fig:rar_train}
    \end{figure}

\begin{algorithm}[t]
\SetAlgoLined
\SetKwInOut{Input}{Input}
\Input{$K$ nodes, minibatch size \textit{b}, encoder $E_{c}$, decoder $D_{c}^{k}$, Loss function, $Loss$, optimizer \textit{SGD}, \# layers $L$} 
{$g_{acc} \longleftarrow 0$}
\For{it = 0, 1, \dots}
{
\For{l = 0, \dots,L}
    {
    {$g_{l} \longleftarrow \nabla Loss + g_{acc}$}\\
    $threshold \longleftarrow min(top\ 0.1\%\ of\ abs(g_{l}))$\\
    $mask \longleftarrow  abs(g_{l}) \geq threshold $\\
    $\tilde{g_{l}} \longleftarrow mask \odot g_{l} $\\
    $g_{acc} \longleftarrow g_{acc} + (\neg mask) \odot g_{l}$\\
    $threshold_{inv} \longleftarrow min(top\ 10\%\ of\ abs(\tilde{g}_{l}))$\\
    $mask_{inv} \longleftarrow abs(\tilde{g}_{l})\geq threshold_{inv} $\\
    $\tilde{g}_{l}^{I} \longleftarrow \tilde{g_{l}} \odot mask_{inv}$
    }
    {$\tilde{g}_{k} \longleftarrow concatenate(\tilde{g}_{l})$}\\
    $\tilde{g}^{I}_{k} \longleftarrow concatenate(\tilde{g}_{l}^{I})$\\
    $\tilde{g}^{c} \longleftarrow E_{c}(\tilde{g}_{k})$\\
    $\tilde{g}^{rec} \longleftarrow D_{c}^{k}(\tilde{g}^{c}, \tilde{g}^{I}_k)$
}
\caption{Compression and reconstruction of the gradient with the LGC, for the parameter server communication pattern, on the node \textit{k} (which is produce compressed-common representation)}
\label{code:PS_code}
\end{algorithm}

    \begin{algorithm}[t]
    \SetAlgoLined
    \SetKwInOut{Input}{Input}
    \SetKwInOut{Output}{Output}
    \Input{$K$ nodes, minibatch size \textit{b}, encoder $E_{c}$, decoder $D_{c}$, Loss function, $Loss$, optimizer \textit{SGD}, \# layers $L$}
    {$g_{acc} \longleftarrow 0$}\\
    \For{it = 0, 1, \dots}
        {
        \For{l = 0, \dots,L}
            {{$g_{l} \longleftarrow \nabla Loss + g_{acc}$}\\
            $threshold \longleftarrow min(top\ 0.1\%\ of\ abs(g_{l}))$\\
             $mask \longleftarrow  abs(g_{l}) \geq threshold $\\
             $\tilde{g_{l}} \longleftarrow mask \odot g_{l} $\\
             $g_{acc} \longleftarrow g_{acc} + (\neg mask) \odot g_{l}$\\
             }
        {$\tilde{g}_{k} \longleftarrow concatenate(\tilde{g}_{l})$}\\
        $\tilde{g}^{c}_{k} \longleftarrow E_{c}(\tilde{g}_{k})$\\
        $\tilde{g}^{avg} \longleftarrow \frac{1}{K}\sum_{k} \tilde{g}^{c}_k(k = 1, \dots, K) $ \\
        $\tilde{g}^{rec} \longleftarrow D_{c}(\tilde{g}^{avg})$}
    \caption{Compression and reconstruction of the gradient with the LGC, for the ring-allreduce communication pattern, on the node \textit{k} }
    \label{code:RAR_code}
    \end{algorithm}

      %%%%%%%%%%%%%%%%%%%%%%
    \subsection{Distributed Training Process}
     \label{sec:TrainingDescription}
    %%%%%%%%%%%%%%%%%%%%%%
    During the first iterations, the weights of a model change very aggressively and thus, the calculated gradients are being rapidly outdated. Any substitution or transformation of the gradients at this stage can be harmful to the performance of the model.
    For this reason, we do not apply the gradient sparsification process and compression at the first iterations of the training.
    
    After a number of iterations, the weights of the model are updated using the gradient vector $\tilde{g}_{k}$ (which has been constructed using the the highest magnitude values, see the~\secref{sec:GradientSparsification} for more details). In parallel, the autoencoder network is trained using the gradient vectors at a particular node. This is either the master node in the \textit{parameter server} communication pattern (see Fig.~\ref{fig:ps_train_autoencoder}) or any selected node in the \textit{ring-allreduce} pattern (see Fig.~\ref{fig:rar_train}). Note that there can be various criteria for selecting the node that is responsible for training the compression network in the \textit{ring-allreduce} pattern, such as energy,  computational capacity, or bandwidth constraints.
    Note that the training of the compression network lasts for a number of iterations (typically, 200-300 iterations depending on the task, see Section~\ref{sec:convergence_autoencoder} for more details) and then it can be used to compress the gradients. To this end, our approach treats the two communication patterns differently.
    
    \subsubsection{LGC - Parameter Server Communication Pattern}
  
    In the \textit{parameter server} communication pattern, the weights of the learned encoder are transferred to one of the worker nodes. 
    It is worth mentioning that in the \textit{parameter server} communication pattern (see~\algref{code:PS_code} for more details), the trained encoder $\textit{E}_c$ of the proposed LGC framework at one given worker node $k$ compresses its (top-magnitude) gradient vector $\tilde{g}_k$ to the representation $\tilde{g}_k^c$, which is in turn transmitted to the master node. In parallel, all worker nodes--including the node mentioned before--apply coarse gradient selection on gradient vectors  $\tilde{g}_k$ with a very aggressive sparsification rate of $0.001\%$, resulting in transmitting the vector $\tilde{g}_k^I\in\mathbb{R}^{0.00001*\mu}$, $k=1,\dots,K$. One can think of  $\tilde{g}_k^{I}$ as the innovation part of $\tilde{g}_k$, which is specific to each worker node, and $\tilde{g}_k^c$ as the common-compressed information shared across all nodes. Therefore, by having only one worker node sharing this information, we can leverage our observations as detailed in~\secref{sec:InfoPlane} (\ie the the common and the innovation gradient  information, see the relevant section for more details).
    
\begin{figure}
        \centering
        \includegraphics[width=0.6\linewidth]{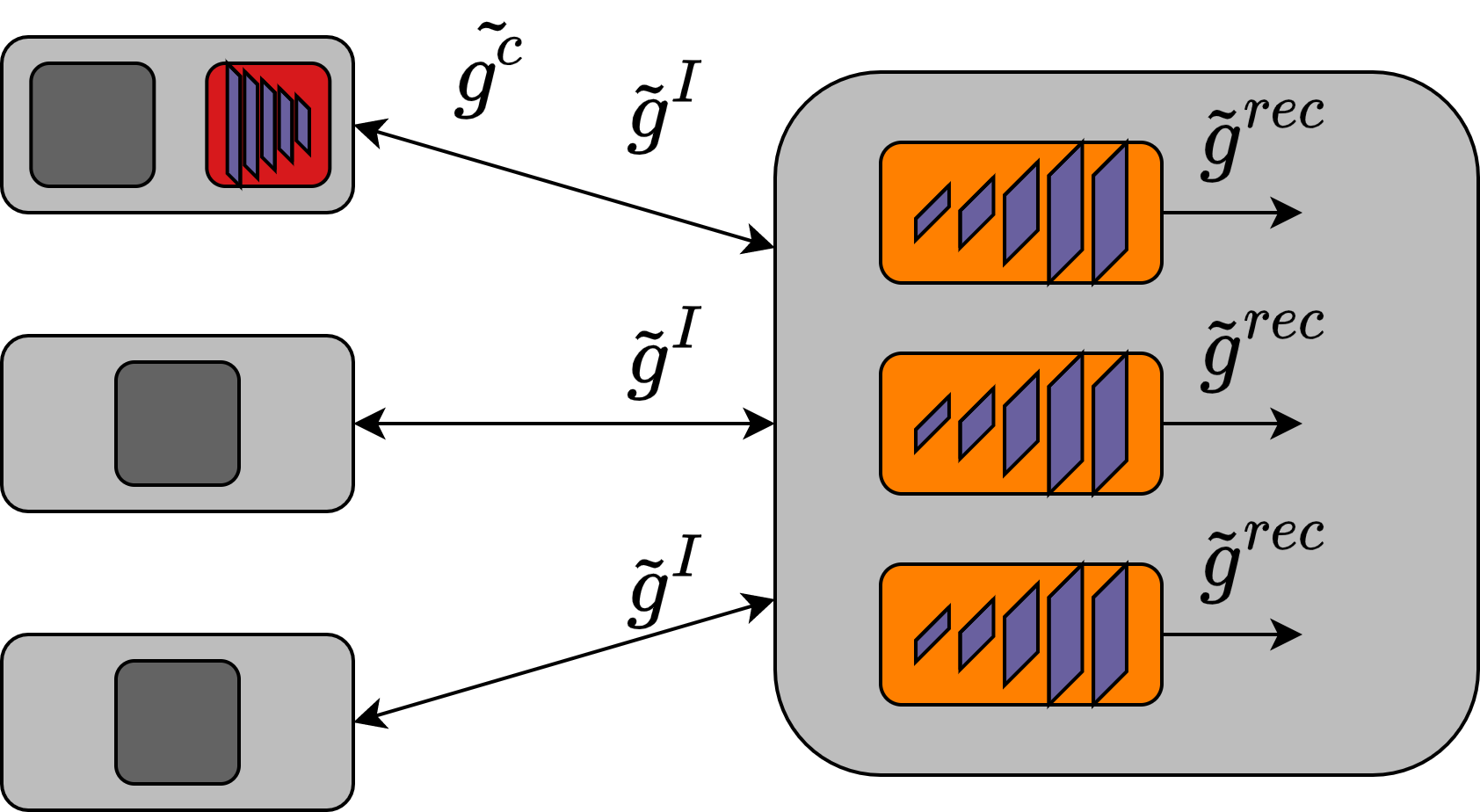}
        \caption{Distributed training within the parameter server communication pattern: $\tilde{g}^{c}$ is compressed-common gradient vector, $\tilde{g}^{I}$ is an innovation gradient vector, and $\tilde{g}^{rec}$ is a reconstructed gradient.}
    \label{fig:PS_training}
    \end{figure}

    At the master node, $\tilde{g}^c$ and $\tilde{g}_k^{I}$ are used to reconstruct the gradient $\tilde{g}_k^{rec}$ with the help of the decoder $D_{c}^{k}$ of the proposed autoencoder (see~\figref{fig:PS_training}), that is,
    \begin{equation}
        \tilde{g}_k^{rec} = D_c(\tilde{g}^c, \tilde{g}_k^{I}).
    \end{equation}
    
    The master node then obtains the aggregated gradient by averaging the reconstructed gradients:
    \begin{equation}
        \tilde{g}^{rec} =  \frac{1}{K} \sum_{k=1}^K \tilde{g}_k^{rec}.
    \end{equation}

    The weight update in the training process is performed in three \revised{consecutive} stages. In the first stage, i.e., during the initial iterations, the weights are updated using the original gradients:
    \begin{equation}
            \label{weight_upd1}
    w_{t}^l = w_{t-1}^l + \lambda \nabla_{t}^{l}
    \end{equation}
    where $w_{t}^l$ is the weight of the layer \textit{l} at iteration \textit{t}. In the second stage, during the training of the compression network, the weights are updated using the top-magnitude gradients:
    \begin{equation}
        \label{weight_upd2}
    w_{t}^l = w_{t-1}^l + \lambda \tilde{g}_{t}^{l}.
    \end{equation}
    In the third stage, the weights are updated with the reconstructed aggregated gradients:
    \begin{equation}
        \label{weight_upd3}
    w_{t}^l = w_{t-1}^l + \lambda \tilde{g}_{t}^{l, rec}.
    \end{equation}
    
    \subsubsection{LGC - Ring-Allreduce Communication Pattern}
    In the \textit{ring-allreduce} pattern, after the training of the autoencoder, the weights of the encoder and the decoder are sent to all other nodes; namely, the $K-1$ nodes, except for the node that the training of the compression network is performed. This communication takes place only once during training and the associated rate, which is negligible, is counted in the total rate (see Section~\ref{sec:Experiments}).

    In the \textit{ring-allreduce} communication pattern (see Algorithm~\ref{code:RAR_code} for more details), the top-magnitude gradient vector $\tilde{g}_k$ at each node~$k=1,\dots,K$, is passed to the encoder $E_{c}$ of the proposed autoencoder, and transformed to the compressed representation:
    \begin{equation}
        \tilde{g}_{k}^c = E_{c}(\tilde{g}_k).
    \end{equation}

     \begin{table}[t]
     
    \caption{Comparison of the proposed LGC framework with the other methods of the data-parallel distributed training. Note that the DGC and Sparse GD refer to the Deep Gradient Compression and Sparse GD methods, respectively. }
    \begin{center}
    \resizebox{0.85\columnwidth}{!}{%
        \begin{tabular}{c|ccc}
        
               & \makecell{LGC} & DGC & Sparse GD \\ 
        \hline\hline
        \thead{Sparse Gradients}  & \thead{\checkmark} &  \thead{\checkmark} &  \thead{\checkmark} \\
        \hline
        \thead{Locally Accumulated Gradients} &  \thead{\checkmark} &  \thead{\checkmark} &  \thead{\checkmark} \\
        \hline
         \thead{Momentum Correlation}  &  \thead{\checkmark} &  \thead{\checkmark} & \thead{\textcolor{red}{-}} \\
         \hline
        \thead{Same Hyperparameters for Distributed\\ and Non-Distributed Training} &  \thead{\checkmark} & \thead{\textcolor{red}{-}} &  \thead{\checkmark} \\ 
         \hline
        \thead{Fixed Sparsification Amount \\during Training} &  \thead{\checkmark} &  \thead{\textcolor{red}{-}}   &  \thead{\checkmark} \\
        \hline
        \thead{Gradient Compression via Autoencoder} &  \thead{\checkmark} & \thead{\textcolor{red}{-}} & \thead{\textcolor{red}{-}} \\
        \hline\hline
        \end{tabular}
        }
        
    \label{table:technique_comparison}
    \end{center}
    
\end{table}

    \begin{figure}
        \centering
        \includegraphics[width=0.6\linewidth]{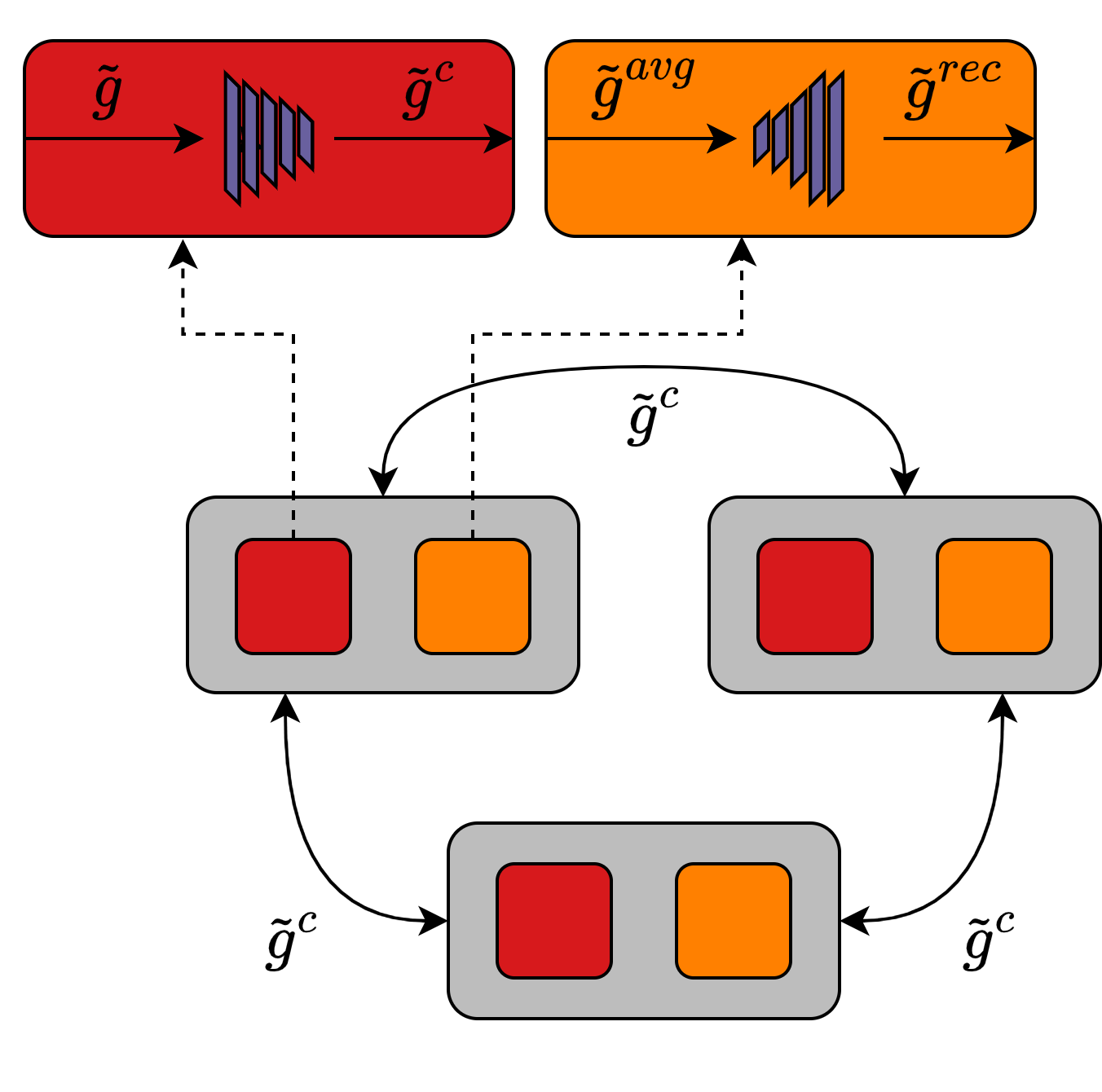}
        \caption{Distributed training within the ring-allreduce communication pattern: $\tilde{g}^{c}$ is a compressed  top-magnitude gradient vector, $\tilde{g}^{avg}$ is a compressed averaged gradient, and  $\tilde{g}^{rec}$ is a final gradient vector.}
        \label{fig:RAR_training}
    \end{figure}
    These compressed representations are exchanged between the nodes, and after this exchange, each node obtains the averaged compressed representation\giannis{:}
    \begin{equation}
    \tilde{g}^{avg} = \frac{1}{K} \sum_{k=1}^K \tilde{g}_k^c.
    \end{equation}
    
    Then, at each node, $\tilde{g}^{avg}$ passed to the decoder part of the proposed autoencoder, delivering the reconstructed aggregated gradient:
    \begin{equation}
    \tilde{g}^{rec} = D_c(\tilde{g}^{avg}).
    \end{equation}

    Finally, the reconstructed aggregated gradients are used to update the weights of the model (see~\figref{fig:RAR_training}).
    In the \textit{ring-allreduce} communication pattern, the weight updates in the training process are also performed in three stages as described by the equations~\eqref{weight_upd1},~\eqref{weight_upd2}, and~\eqref{weight_upd3}.

    In~\tabref{table:technique_comparison}, we compare the proposed methodology LGC with other methods for data-parallel distributed training.
    As we observe, LGC relies on techniques, such as momentum correlation and local gradient accumulation to improve the performance. These techniques has also been used in other works (DGC~\cite{DGC} and Sparse GD~\cite{Strom}). Unlike the aforementioned architectures, our method exploits gradient compression of distributed nodes to reduce the communication rate.
    Furthermore, it is worth mentioning that, in case of LGC, switching from single-node training to distributed training is smooth, as the framework does not require any hyperparameter change in the training flow. \revised{ Moreover, in case of DGC, sparsification rate is changing within a warm-up iterations (\ie strategy of the warm-up is a hyper-parameter introduced by DGC). In contrast, in our case we adopt the pattern of fixed sparsification rate regardless of task (\ie no sparsification at the warm up iterations and sparsification with the highest rate for the rest of the iterations).}

    %%%%%%%%%%%%%%%%%%%%%%
    \section{Experiments}
    \label{sec:Experiments}
    %%%%%%%%%%%%%%%%%%%%%%
    In this section, we present the evaluation of our LGC framework on classification and image-to-image transformation tasks.
     All experiments are performed on a single machine with four GeForce RTX 2080 Ti GPUs and 128 GB of RAM, by emulating more than one node on each GPU. The LGC framework is built on top of Pytorch's~\cite{paszke2017automatic} distributed package. Furthermore, we want to stress that downlink communication is not a focus of this work (like in~\cite{downlink2017terngrad, downlink2018convergence, downlink2019qsparse, DGC}), but it can be inexpensive when the broadcast routine is implemented in a tree-structured manner as in several MPI ( Message Passing Interface) implementations.
    
    \subsection{Experimental Setup}
    Regarding the image classification task, we consider the Resnet50 convolutional neural network trained on the Cifar10~\cite{cifar} and ImageNet~\cite{imagenet} datasets and ResNet101~\cite{resnet} on the Cifar10~\cite{cifar} dataset.
    Regarding the image-to-image transformation task, we consider the PSPNet~\cite{pspnet} model trained on the CamVid~\cite{camvid} semantic segmentation dataset.
    
    The distributed training strategy within the LGC framework is as follows: the model is initially trained without any gradient modification for approximately 200 (depending on the task) iterations. This is because in the first iterations of training, the weights change fast and any transformation at the gradients can reduce the performance of the model (see Section~\ref{sec:sparsification}). Then, the weights are updated using the top-magnitude values of the gradients, and at the same time, the autoencoder is being trained as described in Section~\ref{sec:TheFramework}. This process lasts for another 200 iterations for the image-to-image transformation task and 300 iterations for the classification tasks. The autoencoder is trained with the SGD optimizer using a learning rate of 0.001 and a batch-size of 1. For the remaining iterations---that is, approximately, 89\% and 83\% of the total iterations for the classification and the image-to-image transformation tasks, respectively---the framework performs distributed training with the compressed top-magnitude values of the gradients using the trained autoencoder. Following other studies, e.g.,~\cite{DGC}, the aforementioned process is applied to all layers of the model except for the following two: 
    \begin{enumerate*}[label=\textit{(\roman*)}]
    \item the first layer for which the update of the weights is performed using the original gradients; and
    \item the last layer of the network (fully-connected layer in the case of the classification tasks and convolutional layer in the case of image-to-image transformation task), where the top-magnitude values of the gradient are selected without further compression (\ie the autoencoder is not used). 
    \end{enumerate*}
    In all layers, where the top-magnitude gradient values are selected, we set the sparsity level $\alpha$ to 0.1\%.
    In all experiments, we report the compression ratio (CR) defined as,
    $$\text{CR} = \text{size}\big(G^\text{original}_{k}) / \text{size}\big(G^\text{compressed}_{k}),$$
    where $G^\text{original}_{k}$ and $G^\text{compressed}_{k}$ are the uncompressed and compressed gradients at the training node $k$, and the $\text{size}(\cdot)$ function computes the size of the gradient tensor in Megabytes. Two compression ratios are reported for the proposed LGC framework under the \textit{parameter server} pattern. The first one refers to the worker node that shares the common-compressed and the innovation gradient component and the second one refers to all other nodes where only the innovation component is being sent.
    In what follows, the Baseline refers to performing distributed training of the model with the original, a.k.a.,  uncompressed, gradients.

     \begin{figure}[t]
        \centering
        \includegraphics[width=0.8\linewidth]{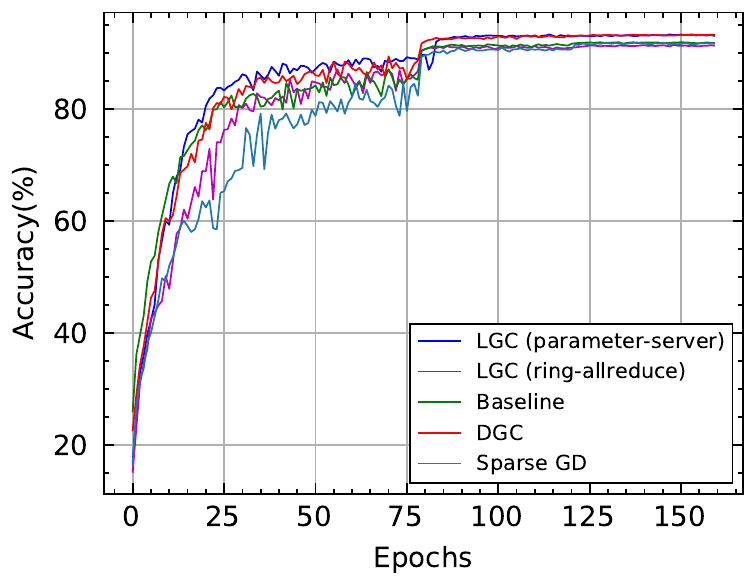}
        \caption{\revised{Comparison of the learning curves in terms of the top1 accuracy  of the Resnet50 model on the image classification task between the two versions of the LGC model (\ie parameter-server and ring-allreduce) and the rest of the examined architectures. Note that the baseline model refers to the distributed training with the uncompressed, non-modified gradients.}}
        \label{fig:resnet50}
    \end{figure}

    \begin{figure}[t]
        \centering
        \includegraphics[width=0.8\linewidth]{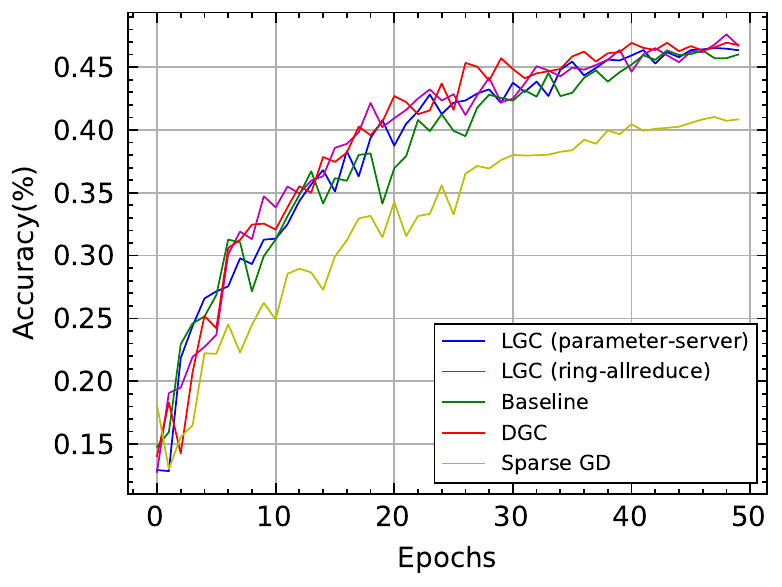}
        \caption{Comparison of the learning curves in terms of the (pixel) accuracy of the  PSPNet model on the semantic segmentation task (CamVid dataset) between the two versions of the LGC model (\ie parameter-server and ring-allreduce) and the rest of the examined architectures. Note that the baseline model refers to thedistributed training with the uncompressed, non-modified gradients.}
        \label{fig:pspnet}
    \end{figure}

    \begin{table}[t]
    \begin{center}
        \caption{\revised{Top 1 accuracy versus the compression ratio in distributed training of ResNet50 on ImageNet. Information is indicating the cumulative amount of information being sent during the whole training. In the case of parameter server, the first number in compression ratio is indicating the metrics of the node, which is sending both compressed common and innovation component, and the second number is for the rest of the nodes. }}
        \resizebox{\columnwidth}{!}{%
    \begin{tabular}{ c||c|l|l} 
    \thickhline
    \thead{Training \\Method} & \thead{Top 1 \\ Accuracy} & \thead{Compression \\ Ratio} & \thead{Information} \\ 
     \thickhline
        Baseline  & 75.98\% & 1$\times$ & 351TB  \\
        \hline
        LGC (parameter server) & 75.88\% & \textbf{386/2800$\times$} & \textbf{0.4TB} \\
        \hline
        LGC (ring-allreduce) & 75.91\% & 202$\times$ & 1.9TB \\
         \hline
        ScaleCom & 75.98\% & 96$\times$ & 3.6TB \\
        \hline
        DGC & \textbf{76.01\%} & 277$\times$ & 1.2TB \\
        \hline
        Sparse GD & 75.54\% & 277$\times$ & 1.2TB \\
         \thickhline
        \end{tabular} 
        }
        \label{table:imagenet}
    \end{center}
\end{table}

\begin{table}[t]
    \begin{center}
        \caption{\revised{Duration of the one iteration (in seconds) of the distributed training for each of three phases of gradient updates on the distributed training of ResNet50 on ImageNet with eight nodes. \textbf{Full update} refers to the update with the uncompressed gradients. \textbf{Top-k update} refers to the phase of the updates with the top-k values of the gradients, and iterations when the autoencoder is trained. \textbf{Compressed update} relates to the updates with the proposed autoencoder. }}
        \resizebox{\columnwidth}{!}{%
    \begin{tabular}{ c||c|l} 
    \thickhline
    \thead{Phase} & \thead{LGC \\ parameter server} & \thead{LGC \\ ring-allreduce}\\
         \thickhline
        Full update  & 1sec & 1sec \\
        \hline
        Top-k update & 1.6sec & 0.9sec \\
        \hline
        Compressed update & 0.6sec & 0.4sec \\
         \thickhline
        \end{tabular} 
        }
        \label{table:latency}
    \end{center}
\end{table}
    
    \subsection{ImageNet Classification}
    \label{sec:imagenet_classification}
    
    \revised{In the first set of experiments, we conduct distributed training of ResNet50 on ImageNet~\cite{imagenet} and assess the achieved accuracy and speedup versus the gradient compression ratio.}
    
    \revised{ImageNet is a large-scale image classification dataset with over $1.2M$ training and $50K$ validation images belonging to $1000$ classes.
    We used the following settings in our experiments on ImageNet: the SGD optimizer with a momentum of $0.9$, a weight decay of $1e-4$, and an initial learning rate of $0.1$ that decays by $10$ every $30$ epochs. Furthermore, we adopt Inception preprocessing with an image size of $224\times224$~\cite{inceptionv4} pixels and a batch size of $256$. The reported results are single-crop performance evaluations on the ImageNet validation set.} 
    
    \revised{For the experimental evaluation of our LGC framework, we performed distributed training on eight nodes (by simulating two nodes on each GPU). Table~\ref{table:imagenet} depicts the classification accuracy versus compression ratio and the total amount of the gradient information (in TBs) sent from all nodes per iteration for the LGC framework and alternative state-of-the-art distributed training methods, namely, DGC~\cite{DGC} (our implementation), ScaleCom~\cite{chen2020scalecom} and Sparse GD~\cite{Strom} (our implementation). The results show that the proposed framework is able to achieve a $386\times$ compression of the gradients without loss of accuracy in the case of parameter-server communication pattern and $202\times$ in the case of ring-allreduce, compared with the baseline of distributed training with the uncompressed, non-modified gradients. Moreover, in the case of the parameter-server communication pattern our results are outperforming the state-of-the-art methods of ScaleCom and DGC, where achieved compressions are $96\times$ and $277\times$, respectively. Furthermore, In Table ~\ref{table:imagenet}, reported total amount of gradient information transferred during the entire training in the case of the LGC, includes updates with the original and the top-k gradients during the first two training phases, as described in Section~\ref{sec:TheFramework}. The total amount is lower by $794\times$ for the parameter-server communication pattern and by $184\times$ for the ring-allreduce compared with the baseline training. Even though the gradient compression ration in ring-allreduce pattern is lower than that in the parameter-server pattern, the gain in the speedup is much higher. For the parameters-server communication pattern, the LGC achieved $1.7\times$ speedup and for the ring-allreduce, $2.56\times$. This is due to that autoencoder latency in ring-allreduce is lower than that of the parameter-server communication pattern; the autoencoder in the latter setting performs a top-k selection for the construction of the innovation gradient. Specifically, the average inference time at the encoder is 0.007 ms and 0.01 ms for the ring-allreduce and the parameter server pattern, respectively, and the average inference time at the decoder is 1 ms in both patterns. Moreover in Table~\ref{table:latency} we are presenting the duration of one iteration for each type of gradient update that we are using within a distributed training.}
    
    \subsection{Cifar10 Classification}
    The Cifar10~\cite{cifar} dataset consists of 50,000 training and 10,000 validation images (with resolution $32\times32$ pixels) from 10 different classes. The distributed training of Resnet50 and Resnet101, on the Cifar10 dataset, is performed on 2 and 4 nodes, respectively. The training procedure and the hyperparameter selection for the models follows that in~\cite{resnet}. \Figref{fig:resnet50} depicts the evolution of the classification accuracy with respect to the number of epochs for the ResNet50 model trained on Cifar10 using the proposed LGC framework. 
    We compare the performance of LGC, under the parameter-server and ring-allreduce communication patterns, against the baseline, DGC~\cite{DGC} and Sparse GD~\cite{Strom}. Furthermore,~\tabref{table:all_experiments} reports the Resnet50 top-1 classification accuracy, the total amount of gradient information sent from all nodes per iteration, and the compression ratio for each framework. The results show that our framework achieves comparable (for the ring-allreduce communication pattern) or even higher classification accuracy (for the parameter-server communication pattern) compared to the baseline approach while reducing by 5709$\times$ and 3193$\times$ the gradient information per iteration in the parameter-server and the ring-allreduce communication patterns, respectively.
    
    Table~\ref{table:all_experiments} presents the performance of the proposed LGC framework (in the parameter server and in the ring-allreduce  communication patterns) when training the ResNet101 model on the Cifar10 dataset using four nodes. Again, we compare LGC against the baseline and DGC~\cite{DGC}. 
    The results show that our approach can drastically reduce the gradient rate compared to the baseline while incurring a small loss in the model performance. Specifically, in the parameter server setup, our approach reduces the size of the gradient information sent at each iteration and per node from 170MB to 0.021MB, while incurring a $0.18\%$ loss in accuracy compared to distributed training with the uncompressed, non-modified gradients. It is also worth noting that our approach reduces significantly, by almost an order of magnitude in the parameter server communication pattern, the gradient rate compared to the state-of-the-art DGC~\cite{DGC}.
    
    \subsection{CamVid Semantic Segmentation}
     CamVid~\cite{camvid} is a semantic segmentation dataset which consists of 32 different classes. The dataset contains 701 images with a spatial resolution of $720\times960$ pixels.
     On the CamVid dataset, we performed training of the PSPNet~\ref{fig:pspnet} model on 2 nodes, with a batch size of 12, using momentum SGD. The network trained with random crops of size $473\times473$, on which augmentations in the form of scaling and rotation are applied.
     
     Table~\ref{table:all_experiments} reports the results obtained with training the PSPNet model on two nodes to address the semantic segmentation task. The results illustrate that our method reduces the size of the gradient information sent from each node at each iteration by a fraction of 450$\times$--700$\times$, depending on the communication setup, without any reduction in terms of performance (\ie pixel accuracy). In effect, the proposed method leads to the same (for the parameter server setup) or even higher (for the ring-allreduce setup) pixel accuracy compared to the baseline method (\ie a PSPNet model without distributed training). Moreover, the compression ratio--and the pixel accuracy (in case of the ring-allreduce setup)--of our LGC model is higher than that obtained with DGC. Fig.~\ref{fig:pspnet} depicts the learning curves of the PSPNet for the various distributed training methods (and the baseline). We notice that the learning curves obtained with LGC and DGC are on par with that of the baseline, whereas the learning curve obtained with the Sparse GD method~\cite{Strom} is significantly lower.

  \begin{table*}[t]
  \begin{center}
      
\caption{ Performance comparison of LGC framework (for parameters server (ps) and ring-allreduce (rar) communication patterns) with the other methods of distributed training. \textbf{Info size} refers to the size of information being sent per forward pass. \textbf{Ratio} corresponds to the compression ratio. In the case of parameter server, the first number is indicating the metrics of the node, which is sending both compressed common representation and innovation, and the second number is for the rest of the nodes. \textbf{Top1/Pixel Acc.} refers to the top1 classification accuracy and the pixel accuracy, correspondingly. }
\begin{tabularx}{\textwidth}{@{} l *{9}{C} @{}}
    \toprule
 
\thead{Training \\ Method}
    &   \mc{\thead{ResNet50 on Cifar10}}
    &   \mc{\thead{ResNet101 on Cifar10}}
    &   \mc{\thead{PSPNet on CamVid}}               \\
    \cmidrule(rl){2-4}\cmidrule(rl){5-7}\cmidrule(rl){8-10}
    &  \thead{Top1}
    &   \thead{Info \\ size}
    &   \thead{Ratio}

    &  \thead{Top1}
    &   \thead{Info \\ size}
    &   \thead{Ratio}

    &  \thead{Pixel \\ Acc.}
    &   \thead{Info \\ size}
    &   \thead{Ratio} \\

 \thickhline
    \hline
    Baseline  & 91.88\% & 102.2MB & 1$\times$ & 93.75\% & 170MB & 1$\times$  & 46.3\% & 120MB & $1\times$\\
    \hline
    Sparse GD & 91.82 & 0.102MB & 1000$\times$ & 92.75 & 0.17MB & 1000$\times$ & 41\% & 0.29MB & 413$\times$ \\
    \hline
    DGC & 93.2\% & 0.102MB & 1000$\times$ & \textbf{93.87\%} & 0.17MB & 1000$\times$  & 46.5\% & 0.29MB & 413$\times$ \\
    \hline
    LGC (RAR)  & 91.4\% & 0.032MB & 3193$\times$ & 93.07\% & 0.074MB & 2297$\times$ & \textbf{47.6\%} & 0.261MB & 459$\times$ \\
    \hline
    LGC (PS)  & \textbf{93.27} \% & \textbf{0.017/ 0.012MB}  & \textbf{5709/ 8616$\times$} & 93.57\% & \textbf{0.021/ 0.01MB} & \textbf{8095/ 17000$\times$}  & 46.3\% & \textbf{0.17/ 0.16MB} & \textbf{693/ 722$\times$} \\
         \thickhline
\end{tabularx}
        \label{table:all_experiments}
          \end{center}

\end{table*}

\begin{figure}   
    \centering
    \subfloat[]{\includegraphics[width=0.4\textwidth]{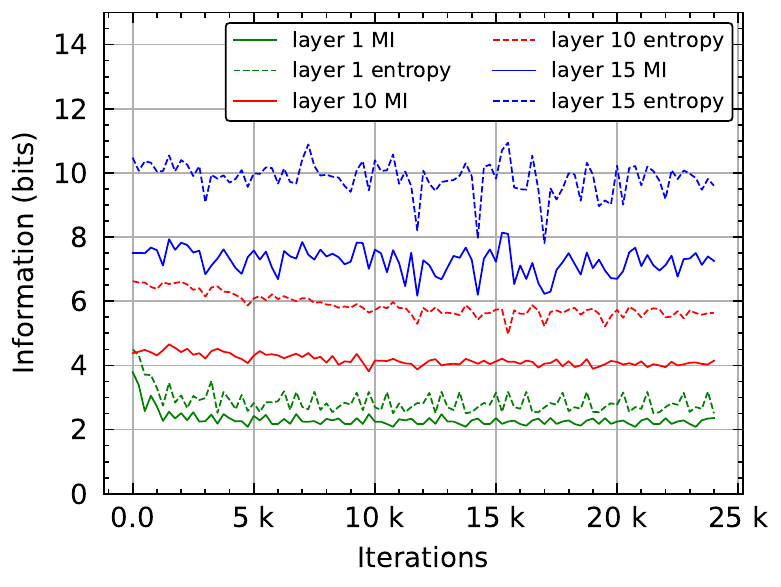}
    \label{fig:mu_vgg}}
    \hfill
    \subfloat[]{\includegraphics[width=0.4\textwidth]{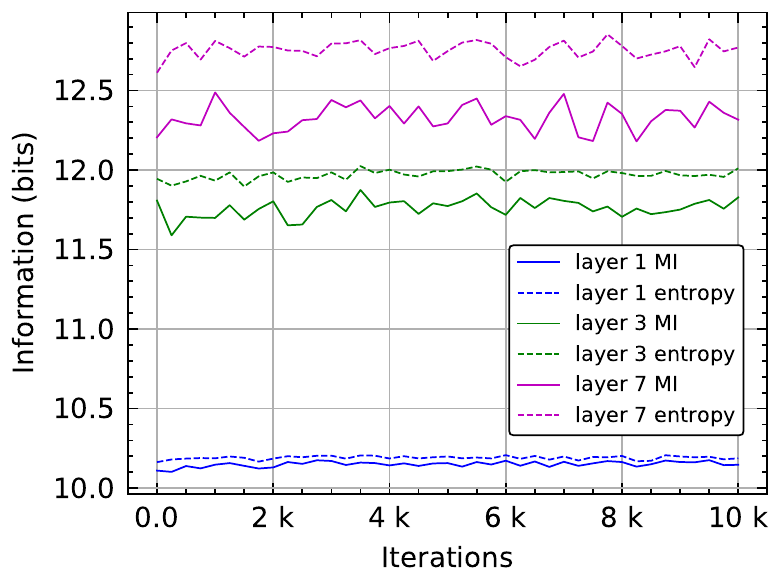}
    \label{fig:mu_convnet5}}
    \caption{\revised{The mutual information (solid lines) and the marginal entropy (dotted lines) between gradient tensors of the same layer on the (a) 3rd and 11th nodes of the VGG11 model, (b) 8th and 10th nodes of the ConvNet5 model, through the training iterations of distributed training.}}
\label{fig:mu_convnet_vgg}
\end{figure}

\subsection{Information Plane in the Large Scale Distributed Training}
\label{sec:info_plane_advanced}
\revised{In order to extend the findings presented in~\secref{sec:InfoPlane} (\ie about the statistical dependencies among the gradient tensors in the distributed training methodology), we conduct two additional experiments in the more complex case that more than two computing nodes are available.
First, we analyze the dependencies among the gradient tensors within the distributed training context of the VGG11 model when 16 nodes are available, and in the custom ConvNet5 model when 22 nodes are available.}

\revised{\noindent \textbf{VGG11}: To conduct our additional distributed training experiment we use a modified version of the VGG16~\cite{vgg} neural network, namely VGG11, comprising 11 convolutional layers. Each convolution is followed by a ReLU nonlinearity. Similar to the original implementation of VGG16, we also adopt a max-pooling operation as a method of spatial dimensionality reduction.
The distributed training was performed on 16 nodes using the Food101~\cite{bossard14food} dataset. Food-101 consists of $75,750$ training and $25,250$ testing images from $101$ different classes. We trained our model for $25$K iterations with a cumulative batch size of $128$, learning rate of $0.001$, using the SGD optimizer.
To analyze the statistical dependencies between the gradient tensors, at each iteration of the training, we construct pairs of gradients of different nodes and calculate the mutual information and the entropy of each gradient pair.
Figure~\ref{fig:mu_vgg} illustrates the mutual information between randomly picked pairs of gradients. As in the case of the ResNet50 (see Section~\ref{sec:InfoPlane}), we observe that a large part of the average information content (\ie the entropy) contained in the gradient tensor of the layer at each iteration is common for both nodes.}

\revised{\noindent \textbf{ConvNet5}: For our second experiment, we construct a convolutional neural network with five convolutional layers, namely ConvNet5. After each convolution, batch normalization~\cite{ioffe:15} and ReLU nonlinearity are added. The model trained on the Tiny ImageNet~\cite{Le2015Tinyimagenet} dataset. It consists of $100,000$ training and $10,000$ validation images from $200$ classes. The distributed training is performed on 22 nodes for $14$k iterations with the cumulative batch size of $128$, a learning rate of $0.001$, and the SGD optimizer. To calculate the mutual information and the entropy, we followed the same procedure as for the VGG11 model. Empirical results (Fig.~\ref{fig:mu_convnet5}) suggest that also in this case the mutual information is high, indicating that there is a considerable amount of information that can be exploited among the gradient tensors. This is also the reason that explains the benefit of our architecture compared to the rest of the distributed training approaches.}

\begin{figure}[t]   
    \centering
    \subfloat[]{\includegraphics[width=0.4\textwidth]{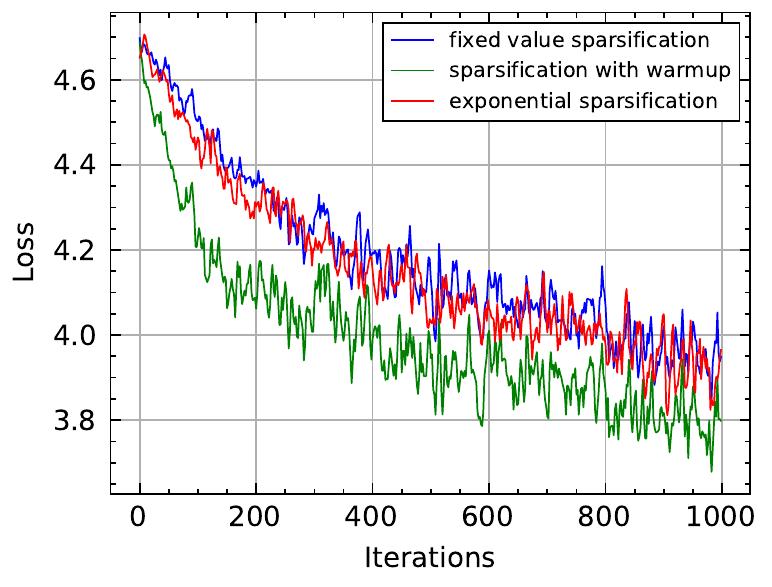}
    \label{fig:sp_custom}}
    \hfill
    \subfloat[]{\includegraphics[width=0.4\textwidth]{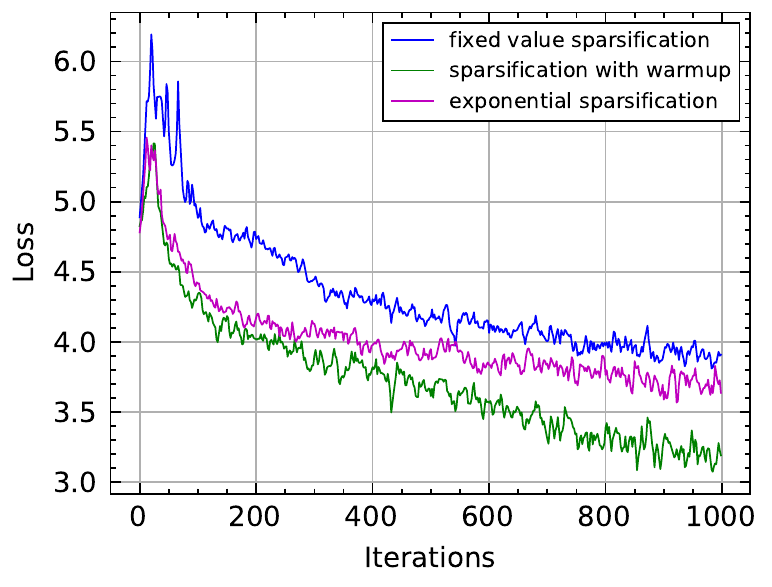}
    \label{fig:sp_resnet}}
    \caption{\revised{Loss of the (a) ConvNet5 (b) ResNet50 within a distributed training for the different types of sparsification strategies: (i) fixed value sparsification refers to the case when sparse updates applied from the first iteration, (ii) exponential sparsification is describing the case when the percentage of sparsification is increasing exponentially during the first iterations, (iii) sparsification with the warmup is describing the case when sparsification is being applied only after the first iterations (e.g. no sparsification at the beginning of the training).}}
\label{fig:sparse_strategy}
\end{figure}

\subsection{Sparsification Strategy}
\label{sec:sparsification}
\revised{To empirically validate the advantage of our choice of sparsification strategy, presented in Section~\ref{sec:TheFramework}, we conduct an experimental comparison between our method and the other methods widely used in distributed training. Specifically, we compare three approaches: (i) the method of exponential increase used in the DGC (e.g. exponentially increase the gradient sparsity from a relatively small value to the final value), (ii) the technique of applying fixed value sparsification within the whole training (e.g. a fixed value of gradient sparsity from the first iteration and till the last) used in~\cite{Strom, QSGD, chen2020scalecom}, (iii) the sparsification with the warmup used in our method (e.g. no sparsification at the first iterations and fixed value gradient sparsification for the rest of the iterations).  The experiments performed on the two types of neural networks - on the compact neural network ConvNet5 (see the description in paragraph~\ref{sec:info_plane_advanced} ) and the model with the large number of parameters, ResNet50. 
Experimental results presented in the Figure~\ref{fig:sparse_strategy} illustrate the advantage of our method. Notably, both in the case of a fixed value and exponential sparsification, we can notice that the loss is decreasing visibly slower, which can lead to the poor convergence of the model. In contrast, in our case, by performing updates with the original gradients within first iterations and switching to the sparse updates after, we can reach a faster decrease in the loss and better convergence. Our extensive experimental results suggest that independent of the model, 200 iterations for the updates with the original gradients are enough to avoid any further mislead in the optimization process caused by gradient sparsification.}

\begin{figure}[t]   
    \centering
    \subfloat[]{\includegraphics[width=0.4\textwidth]{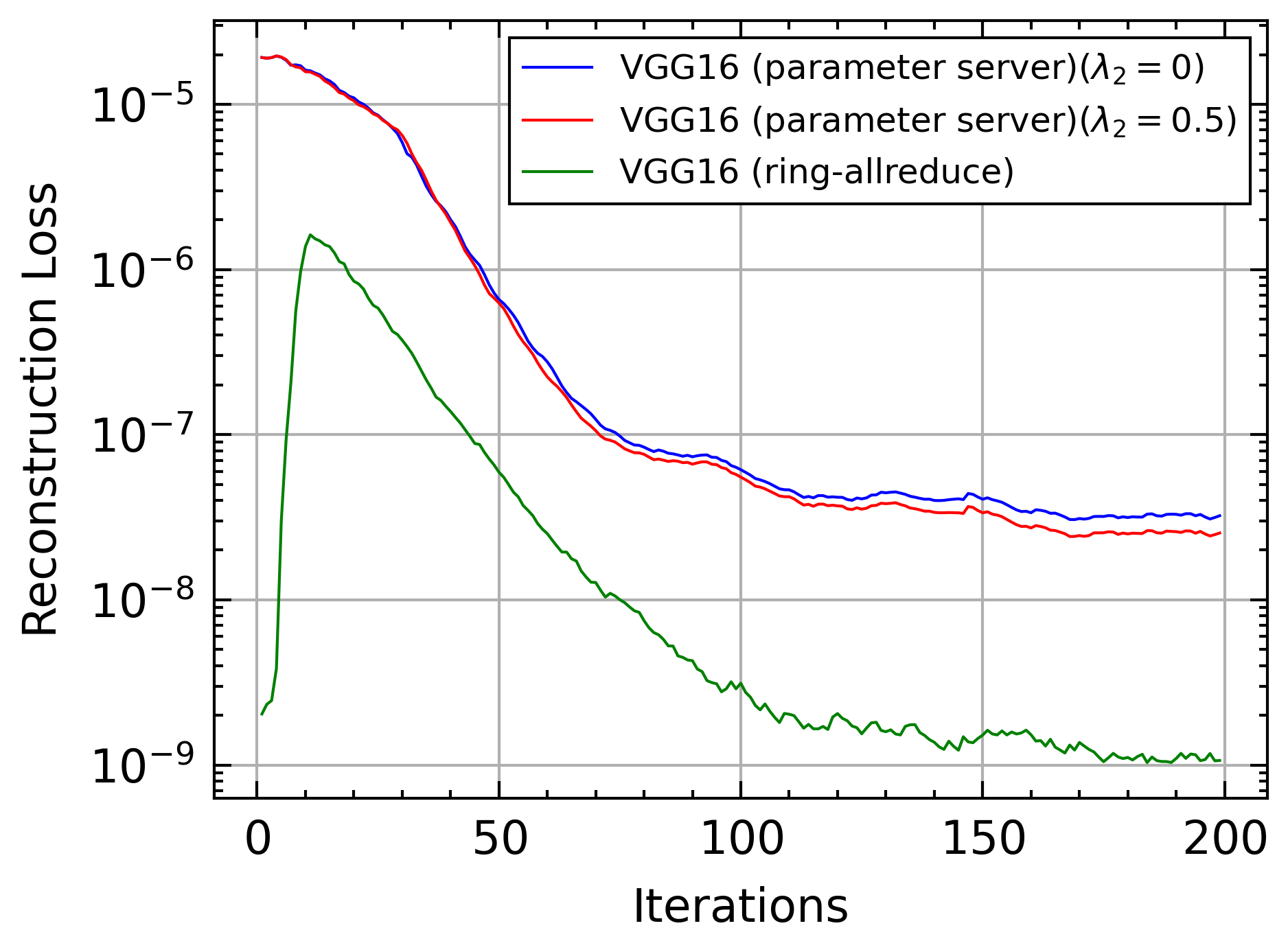}
    \label{fig:conv_vgg}}
    \hfill
    \subfloat[]{\includegraphics[width=0.4\textwidth]{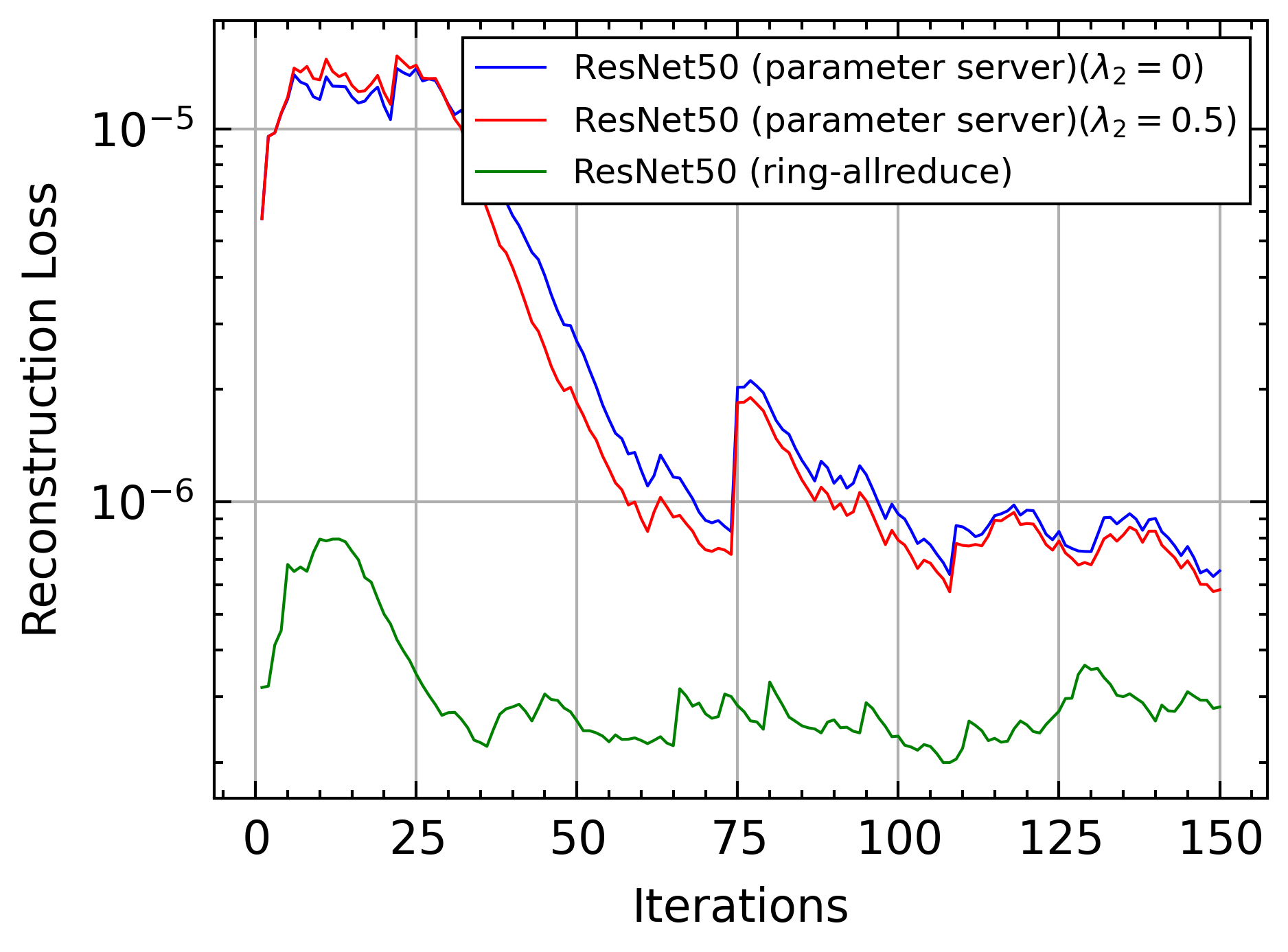}
    \label{fig:conv_resnet}}
    \caption{\revised{Reconstruction loss of the proposed autoencoders within a distributed training of (a) VGG16 (b) ResNet50, where $\lambda_{2}$ is a coefficient of the similarity loss (see Equation~\ref{eq:loss_ps}). $\lambda_{2}=0$ and $\lambda_{2}=0.5$ refers to the cases of training the autoencoder without and with the similarity loss.}}
\label{fig:convergence_autoencoder}
\end{figure}

\subsection{Convergance of the Autoencoders}
\label{sec:convergence_autoencoder}
\revised{We also examined the convergence of our autoencoders (one for the case of parameter-server communication pattern and one for the ring-allreduce) depending on the number of distributed nodes and the primary model (\ie the model being trained in a distributed manner). We conduct experiments on the VGG16~\cite{vgg} model trained on the Tiny ImageNet dataset~\cite{Le2015Tinyimagenet} with the 32 distributed nodes and on the ResNet50~\cite{resnet} model on the Cifar100~\cite{cifar} dataset with the 24 nodes.}

\revised{Results presented in the Figure~\ref{fig:convergence_autoencoder} illustrate the behaviour of the reconstruction loss within training iterations. For the parameter server scenario, we have randomly selected the reconstruction loss of 5th and 32nd nodes in case of the ResNet50 and VGG16 training correspondingly. For all cases, we can observe good convergence of the autoencoders. For the parameter server scenario we have also examined the impact of similarity loss(see Section~\ref{sec:ParameterAutoencoder}) on the reconstruction error. Results presented in Figure~\ref{fig:convergence_autoencoder} suggests that similarity loss(\ie $lambda_{2}=0.5$) helps to reconstruct the gradients better.}

\section{Conclusion}
\label{sec:Conclusion}

In this paper, we have introduced a novel method for data-parallel
distributed training of deep neural networks. It was shown empirically
that the method is able to compress the gradients by $99.99\%$ on image classification tasks, without any reduction in terms of the accuracy or rate of convergence. This was made possible by exploring the correlations between the gradients of different nodes within the scope of distributed training and designing distributed autoencoders for gradient compression.
To the best of our knowledge, the compression rate achieved on the image classification task is the highest ever reported among the other methods on the same network and the same dataset. \revised{Moreover, our LGC framework can provide up to $794\times$ reduction in the total number of bits, being transferred within a distributed training of ResNet50 on ImageNet, compared with the baseline distributed training with original uncompressed gradients.}

\bibliographystyle{IEEEtran}
\bibliography{ldgc_jrnl}

% that's all folks
\end{document}